\documentclass[final,5p,times,twocolumn]{elsarticle}
\usepackage{amsthm,amsmath,amssymb}
\usepackage{mathrsfs}
\usepackage{booktabs}
\usepackage{multirow}
\usepackage{multicol}
\usepackage{color}
\usepackage{hyperref}
\usepackage{float}
\usepackage{subfig}
\usepackage[table]{xcolor} % 加载xcolor包，并且指定table选项
\usepackage{colortbl} % 允许为表格行和列定义颜色
\usepackage{rotating} % 导入 rotating 包
\usepackage{graphicx}
\usepackage[switch]{lineno}
\usepackage[ruled, linesnumbered]{algorithm2e}

\definecolor{LightCyan}{rgb}{0.88,1,1}

\journal{Elsevier}

\begin{document}

\begin{frontmatter}

\title{GPT-4V with Emotion: A Zero-shot Benchmark for Generalized Emotion Recognition}

\author[label1]{Zheng Lian}
\author[label1,label2]{Licai Sun}
\author[label1,label2]{Haiyang Sun}
\author[label3]{Kang Chen}
\author[label1,label2]{Zhuofan Wen}
\author[label1,label2]{Hao Gu}
\author[label1,label2]{Bin Liu \corref{cor1}}
\author[label4,label5]{Jianhua Tao \corref{cor1}}

\affiliation[label1]{
	organization={Institute of Automation, Chinese Academy of Sciences},
	city={Beijing},
	country={China}
}

\affiliation[label2]{
	organization={School of Artificial Intelligence, University of Chinese Academy of Sciences},
	city={Beijing},
	country={China}
}

\affiliation[label3]{
	organization={Software Engineering Institute, Peking University},
	city={Beijing},
	country={China}
}

\affiliation[label4]{
	organization={Department of Automation, Tsinghua University},
	city={Beijing},
	country={China}
}

\affiliation[label5]{
	organization={Beijing National Research Center for Information Science and Technology, Tsinghua University},
	city={Beijing},
	country={China}
}

\cortext[cor1]{Corresponding author}

\begin{abstract}
Recently, GPT-4 with Vision (GPT-4V) has demonstrated remarkable visual capabilities across various tasks, but its performance in emotion recognition has not been fully evaluated. To bridge this gap, we present the quantitative evaluation results of GPT-4V on 21 benchmark datasets covering 6 tasks: \emph{visual sentiment analysis}, \emph{tweet sentiment analysis}, \emph{micro-expression recognition}, \emph{facial emotion recognition}, \emph{dynamic facial emotion recognition}, and \emph{multimodal emotion recognition}. This paper collectively refers to these tasks as ``Generalized Emotion Recognition (GER)''. Through experimental analysis, we observe that GPT-4V exhibits strong visual understanding capabilities in GER tasks. Meanwhile, GPT-4V shows the ability to integrate multimodal clues and exploit temporal information, which is also critical for emotion recognition. However, it's worth noting that GPT-4V is primarily designed for general domains and cannot recognize micro-expressions that require specialized knowledge. To the best of our knowledge, this paper provides the first quantitative assessment of GPT-4V for GER tasks. We have open-sourced the code and encourage subsequent researchers to broaden the evaluation scope by including more tasks and datasets. Our code and evaluation results are available at: \textcolor[rgb]{0.93,0.0,0.47}{https://github.com/zeroQiaoba/gpt4v-emotion}.
\end{abstract}

\begin{keyword}
Generalized Emotion Recognition (GER); GPT-4 with Vision (GPT-4V); Zero-shot Benchmark; Multimodal Fusion; Temporal Modeling.
\end{keyword}

\end{frontmatter}

% \linenumbers

\section{Introduction}
\label{sec1}
Emotion has attracted increasing attention from researchers due to its importance in human-computer interaction. Current research on emotion recognition primarily centers on two aspects: one aims to identify the emotions evoked by stimuli, essentially predicting how viewers might feel after seeing these stimuli \cite{you2017visual}; the other aims to analyze the emotions conveyed by humans in various ways \cite{lian2022smin}. We collectively refer to these tasks as ``Generalized Emotion Recognition (GER)''.

Emotions are associated with lexical, visual, and acoustic information. Among them, visual information (such as colorfulness, brightness, facial expression, human action, etc) contains rich emotion-related content \cite{yang2023emoset}. To enhance visual understanding capabilities, researchers have proposed various algorithms and achieved noteworthy progress. With the development of deep learning, current research has shifted from handcrafted features to deep neural networks. Recently, GPT-4V has showcased impressive visual understanding capabilities across various tasks and domains. This leads to a question: can GPT-4V solve the GER problem to some extent? If so, what is the future direction following the emergence of GPT-4V?

In September 2023, GPT-4V was integrated into ChatGPT, triggering a series of user reports aimed at investigating its visual capabilities \cite{yang2023dawn}. At that time, OpenAI had not yet released the API service, and users could only manually upload test samples to the web service. Due to the high manual effort required, these reports generally involved a limited number of samples per task and merely provided qualitative insights into GPT-4V. In November 2023, OpenAI released the API, but it was initially limited to 100 requests per day. It remained difficult to evaluate GPT-4V against state-of-the-art systems on benchmark datasets. Recently, OpenAI has increased the daily limit, allowing us to conduct more comprehensive evaluations.

\begin{figure}[t]
	\centering
	\includegraphics[width=\linewidth]{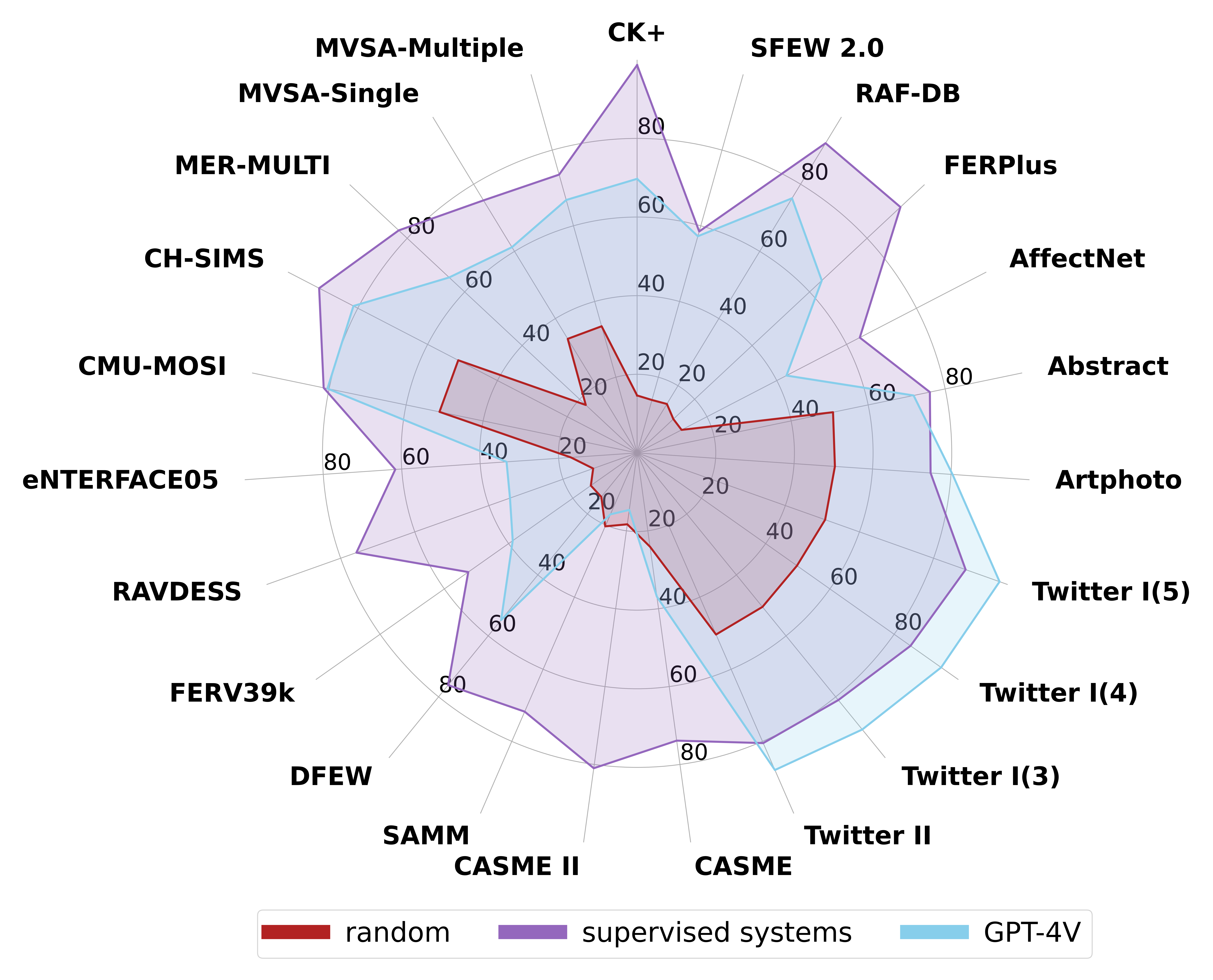}
	\caption{Performance of different methods on GER tasks. Here, ``random'' is a heuristic baseline that randomly selects labels from candidate categories.}
	\label{Figure1}
\end{figure}

In this paper, we provide quantitative evaluation results of GPT-4V on GER tasks, covering visual sentiment analysis \cite{ortis2020survey}, tweet sentiment analysis \cite{niu2016sentiment}, micro-expression recognition \cite{li2022deep}, facial emotion recognition \cite{li2020deep}, dynamic facial emotion recognition \cite{wang2022ferv39k}, and multimodal emotion recognition \cite{lian2024merbench}. Figure \ref{Figure1} shows the overall results of GPT-4V. We also report results of random guessing (a baseline that randomly selects labels from candidate classes) and supervised systems. To fairly compare different methods, we conduct experiments on benchmark datasets and use the same evaluation metric. Overall, GPT-4V outperforms random guessing but still lags behind supervised systems. To figure out the reason, we further conduct a comprehensive analysis of GPT-4V's multi-faceted capabilities, including multimodal fusion, temporal modeling, robustness, stability, etc. We hope this paper can inspire subsequent researchers and shed light on tasks that GPT-4V can effectively address and those that require further exploration. The main contribution of this paper can be summarized as follows:

\begin{itemize}
	
	\item To the best of our knowledge, this is the first work to quantitatively assess GPT-4V's performance in GER tasks.
	
	\item Our evaluation results indicate that GPT-4V is primarily designed for general-purpose domains. It performs poorly in micro-expression recognition that requires specialized knowledge. Meanwhile, it can integrate multimodal clues and capture temporal information during inference.
	
	\item This paper can provide guidance on potential future directions following the emergence of GPT-4V. It can also serve as a zero-shot benchmark for subsequent research.
	
\end{itemize}

The remainder of this paper is organized as follows: In Section \ref{sec2}, we briefly review recent works. In Sections \ref{sec3} and \ref{sec4}, we provide an overview of GER tasks and datasets, along with a detailed description of our GPT-4V calling strategy. In Section \ref{sec5}, we report results and conduct an in-depth analysis of GPT-4V's multi-faceted capabilities. In Sections \ref{sec6} and \ref{sec7}, we summarize the main challenges and limitations of GPT-4, discuss potential directions for future research, and conclude the entire paper.

\section{Related Works}
\label{sec2}

\subsection{Generalized Emotion Recognition}
GER covers a variety of tasks and this paper focuses on 6 representative tasks. In this section, we detail the similarities and distinctions among them. \emph{Visual sentiment analysis} \cite{ortis2020survey} explores the impact of image stimuli on people's emotions, while \emph{tweet sentiment analysis} \cite{niu2016sentiment} aims to determine how individuals feel when posting visual and textual content on social networks. In these tasks, the image is not necessarily human-centric. In contrast, \emph{facial emotion recognition} \cite{li2020deep} aims to identify emotions in human-centric images, particularly focusing on observable macro-expressions. Besides macro-expressions, there is another type of expression known as micro-expressions, which are characterized by short duration, low intensity, and sparse facial action units. The task of identifying such expressions is termed \emph{micro-expression recognition} \cite{li2022deep}. All of the aforementioned tasks identify emotions from static images. \emph{Dynamic facial emotion recognition} \cite{wang2022ferv39k} extends the analysis from static images to image sequences, requiring the use of temporal information. Furthermore, emotions can be conveyed through multiple modalities. Therefore, \emph{multimodal emotion recognition} \cite{lian2024merbench} is introduced to integrate multimodal clues for a more comprehensive understanding of emotions.

\subsection{Unimodal Emotion Recognition}
Emotions can be conveyed through multiple modalities, such as images, text, and audio. Since the early 1970s, Ekman et al. \cite{ekman1978facial, ekman1994strong} have conducted prior studies on facial expressions and developed the Facial Action Coding System (FACS) to code facial expressions. In this system, the face can be described as a set of action units, each associated with a muscular basis. Besides facial expressions, emotions can also be conveyed through text and audio, which carry important communicative messages. Text and audio are highly correlated. On the one hand, we can recognize the text from audio. On the other hand, if we ignore the manner in which the text is spoken, we might misunderstand the meaning of the text \cite{sebe2005multimodal}. With the development of deep learning, researchers have proposed various effective frameworks for emotion recognition \cite{lian2021ctnet}. These models are usually task-specific, i.e., we need to train on emotion corpora. In contrast, this paper uses GPT-4V, which can recognize emotions in a zero-shot manner without further training.

\subsection{Multimodal Emotion Recognition}
Humans express emotions in a complementary manner across various modalities. To better understand emotions, we need to integrate multimodal clues. According to the fusion location, current methods can be roughly divided into three categories: feature-level fusion, model-level fusion, and decision-level fusion \cite{wu2014survey}. Specifically, feature-level fusion concatenates unimodal features at the input level, while decision-level fusion combines unimodal decision results. Despite their promising performance, these methods ignore interactions and correlations between different modalities. Therefore, researchers propose model-level fusion to capture multimodal dependencies. Recently, Lian et al. \cite{lian2024merbench} built MERBench and conducted a systematic analysis of different model-level fusion strategies. Unlike the above methods, GPT-4V achieves multimodal fusion through prompts. More details can be found in Table \ref{Table2}.

\begin{table*}[t]
	\centering
	\renewcommand\tabcolsep{3.6pt}
	\caption{Statistics for different datasets. For each dataset, we evaluate performance on the official test set and select the most common evaluation metric. In this table, ACC, WAR, and WAF represent accuracy, weighted average recall, and weighted average F-score, respectively.}
	\label{Table1}
	\scalebox{0.8}{
		\begin{tabular}{c|lccl}
			\hline
			Task (Modality) & Dataset & \# Test samples & Metric & Labels\\
			\hline
			
			\multirow{4}{*}{{\begin{tabular}[c]{@{}c@{}}Visual Sentiment Analysis \\ (Image)\end{tabular}}} 
			&Twitter I \cite{you2015robust} & 1,269 & ACC & \textcolor[rgb]{0.93,0.0,0.47}{positive, negative} \\
			&Twitter II \cite{borth2013large} & 603 & ACC & \textcolor[rgb]{0.93,0.0,0.47}{positive, negative} \\
			&Abstract \cite{you2016building} & 228 & ACC & \textcolor[rgb]{0.93,0.0,0.47}{amusement, {sadness}, {anger}, {fear}, {disgust}, {awe}, {content}, {excitement}} \\
			&ArtPhoto \cite{you2016building} & 806 & ACC & \textcolor[rgb]{0.93,0.0,0.47}{amusement, {sadness}, {anger}, {fear}, {disgust}, {awe}, {content}, {excitement}}\\
			
			\hline
			
			\multirow{2}{*}{{\begin{tabular}[c]{@{}c@{}}Tweet Sentiment Analysis \\ (Image, Text)\end{tabular}}} 
			&MVSA-Single \cite{niu2016sentiment} & 451 & ACC & \textcolor[rgb]{0.93,0.0,0.47}{positive, neutral, negative} \\
			&MVSA-Multiple \cite{niu2016sentiment} & 1,702 & ACC & \textcolor[rgb]{0.93,0.0,0.47}{positive, neutral, negative} \\
			
			\hline
			
			\multirow{3}{*}{{\begin{tabular}[c]{@{}c@{}}Micro-expression Recognition \\ (Image)\end{tabular}}} 
			&CASME \cite{yan2013casme} 	& 195 & ACC & \textcolor[rgb]{0.93,0.0,0.47}{{happiness}, {sadness}, {fear}, {surprise}, {disgust}, {repression}, {contempt}, {tense}} \\
			&CASME II \cite{yan2014casme} & 247 & ACC & \textcolor[rgb]{0.93,0.0,0.47}{{happiness}, {sadness}, {fear}, {surprise}, {disgust}, {repression}, {others}} \\
			&SAMM \cite{davison2016samm} 	& 159 & ACC & \textcolor[rgb]{0.93,0.0,0.47}{{happiness}, {sadness}, {fear}, {surprise}, {disgust}, {contempt}, {anger}, {others}} \\
			
			\hline
			
			\multirow{5}{*}{{\begin{tabular}[c]{@{}c@{}}Facial Emotion Recognition \\ (Image)\end{tabular}}} 
			&CK+ \cite{lucey2010extended} 			& 981 & ACC & \textcolor[rgb]{0.93,0.0,0.47}{{happiness}, {sadness}, {anger}, {fear}, {disgust}, {surprise}, {contempt}} \\
			&SFEW 2.0 \cite{dhall2015video} 		& 436 & ACC & \textcolor[rgb]{0.93,0.0,0.47}{{happiness}, {sadness}, {anger}, {fear}, {disgust}, {surprise}, {neutral}} \\
			&RAF-DB \cite{li2017reliable} 			& 3,068 & ACC & \textcolor[rgb]{0.93,0.0,0.47}{{happiness}, {sadness}, {anger}, {fear}, {disgust}, {surprise}, {neutral}} \\
			&FERPlus \cite{barsoum2016training} 	& 3,589 & ACC & \textcolor[rgb]{0.93,0.0,0.47}{{happiness}, {sadness}, {anger}, {fear}, {disgust}, {surprise}, {neutral}, {contempt}} \\
			&AffectNet \cite{mollahosseini2017affectnet} & 4,000 & ACC & \textcolor[rgb]{0.93,0.0,0.47}{{happiness}, {sadness}, {anger}, {fear}, {disgust}, {surprise}, {neutral}, {contempt}} \\
			
			\hline
			
			\multirow{4}{*}{{\begin{tabular}[c]{@{}c@{}}Dynamic Facial Emotion Recognition \\ (Video)\end{tabular}}} 
			&DFEW (fd1) \cite{jiang2020dfew} 	     & 2,341 & WAR & \textcolor[rgb]{0.93,0.0,0.47}{{happiness}, {sadness}, {anger}, {fear}, {disgust}, {surprise}, {neutral}} \\
			&FERV39k \cite{wang2022ferv39k} 	  	 & 7,847 & WAR & \textcolor[rgb]{0.93,0.0,0.47}{{happiness}, {sadness}, {anger}, {fear}, {disgust}, {surprise}, {neutral}} \\
			&RAVDESS \cite{livingstone2018ryerson}  & 1,440 & WAR & \textcolor[rgb]{0.93,0.0,0.47}{{happiness}, {sadness}, {anger}, {fear}, {disgust}, {surprise}, {neutral}, {calm}} \\
			&eNTERFACE05 \cite{martin2006enterface} & 1,287 & WAR & \textcolor[rgb]{0.93,0.0,0.47}{{happiness}, {sadness}, {anger}, {fear}, {disgust}, {surprise}} \\

			\hline
			
			\multirow{3}{*}{{\begin{tabular}[c]{@{}c@{}}Multimodal Emotion Recognition \\ (Video, Audio, Text)\end{tabular}}} 
			&CMU-MOSI \cite{zadeh2017tensor}   & 686 & WAF & \textcolor[rgb]{0.93,0.0,0.47}{sentiment intensity}\\
			&CH-SIMS \cite{yu2020ch} 			& 457 & WAF & \textcolor[rgb]{0.93,0.0,0.47}{sentiment intensity}\\
			&MER-MULTI \cite{lian2023mer} 		& 411 & WAF & \textcolor[rgb]{0.93,0.0,0.47}{{happiness}, {sadness}, {anger}, {surprise}, {neutral}, {worry}} \\
			
			\hline
		\end{tabular}
	}
\end{table*}

\begin{figure*}[t]
	\centering
	\includegraphics[width=\linewidth]{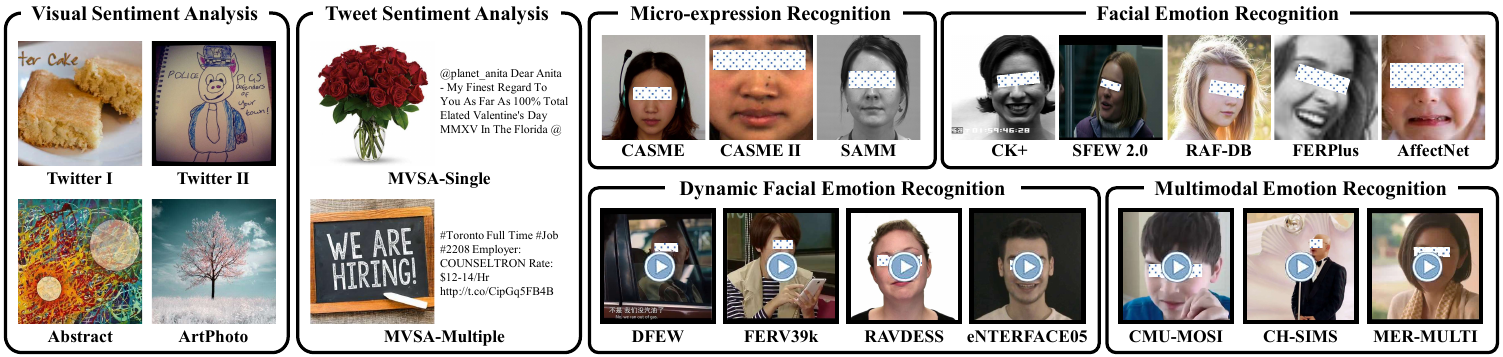}
	\caption{Examples of different datasets. We pixelate faces due to the sensitive nature of human identity.}
	\label{Figure2}
\end{figure*}

\subsection{Multimodal Large Language Model}
Large language models (LLMs) with massive parameters have attracted increasing attention due to their unprecedented performance. In GER tasks, visual information plays an important role in emotion recognition, but LLMs only support lexical inputs. Therefore, we focus on multimodal LLMs (MLLMs) that support visual inputs. Among all MLLMs, MiniGPT-4 \cite{zhu2023minigpt} and LLaVA \cite{liu2023visual} are two representative models. They enhance the visual capability by aligning the frozen visual encoder with LLMs using a simple projection layer, which preserves most of the information encoded in pre-trained models and achieves good performance in visual understanding. This approach is further extended by other works and applies to support more modalities, such as video \cite{li2023videochat}, audio \cite{zhang2023speechgpt}, and depth images \cite{su2023pandagpt}. Currently, GPT-4V is the state-of-the-art MLLM for various tasks \cite{alpaca_eval, lu2024mathvista}. However, its performance in emotion recognition has not been fully studied. Hence, this paper evaluates the performance of GPT-4V on GER tasks, providing a valuable zero-shot benchmark for other MLLMs.

\subsection{Evaluation on GPT-4V}
With the emergence of GPT-4V, a large number of user reports have surfaced to evaluate its visual ability \cite{lin2023mm, wu2023can}. Among them, Yang et al. \cite{yang2023dawn} conducted the most comprehensive assessment, covering coding, captioning, and reasoning abilities. However, they primarily performed qualitative analysis on limited data and did not disclose the performance gap between GPT-4V and supervised systems on benchmark datasets. To bridge this gap, there have been efforts to quantitatively analyze GPT-4V's mathematical reasoning \cite{lu2024mathvista} and visual abilities \cite{wu2023gpt4vis}, but none have specifically explored its emotional understanding capabilities. In this paper, we provide the first quantitative evaluation of GPT-4V on GER tasks. It is worth noting that the emotion is related to humans. However, human identity is treated as sensitive information in GPT-4V, which raises new requirements for the calling strategy. To this end, we also design a calling strategy specifically for GER tasks.

\section{Task Description}
\label{sec3}
In this section, we describe each task and dataset in detail. Table \ref{Table1} summarizes statistics of different datasets. To ensure a fair comparison with supervised systems, we evaluate performance on the official test set and select the most common evaluation metric. Figure \ref{Figure2} shows examples of different datasets, demonstrating the diversity of dataset types. Some datasets are in-the-wild (such as AffectNet), while others are laboratory-controlled (such as CASME and CK+). Meanwhile, there is a disparity in color spaces across distinct datasets. Some datasets use grayscale images (such as CK+), while others use RGB images (such as CASME and AffectNet).

\textbf{Visual Sentiment Analysis} aims to identify the emotions that individuals might evoke from images. We focus on four datasets: Twitter I \cite{you2015robust}, Twitter II \cite{borth2013large}, ArtPhoto \cite{you2016building}, and Abstract \cite{you2016building}. Specifically, Twitter I \cite{you2015robust} and Twitter II \cite{borth2013large} are collected from social websites. Of the two, Twitter I provides raw annotations from 5 Amazon Mechanical Turk workers. Consistent with previous works, we employ three label processing methods, resulting in Twitter I(3), Twitter I(4), and Twitter I(5). Taking Twitter I(4) as an example, it contains all samples where at least 4 workers assign the same label. Different from Twitter I and Twitter II, ArtPhoto \cite{you2016building} contains artistic photographs from a photo-sharing site, while Abstract \cite{you2016building} consists of peer-rated abstract paintings. Both datasets are annotated into eight categories. In line with previous works, we reassign these labels into positive and negative classes and report results for the negative/positive classification task.

\textbf{Tweet Sentiment Analysis} aims to determine how people feel when posting visual and textual content on social networks. We conduct experiments on two benchmark datasets: MVSA-Single \cite{niu2016sentiment} and MVSA-Multiple \cite{niu2016sentiment}. These datasets contain image-text pairs collected from Twitter, and annotators assign one of three sentiments (positive, negative, and neutral) to the image and text respectively. MVSA-Single contains 5,129 image-text pairs labeled by a single annotator. MVSA-Multiple is an extended version of MVSA-Single with more samples and more annotators. For a fair comparison, we adopt the same dataset processing strategy in previous works, and then randomly select 10\% of the samples as the test set \cite{xu2017multisentinet, zhu2022multimodal}.

\textbf{Facial Emotion Recognition} focuses on five benchmark datasets: CK+ \cite{lucey2010extended}, FERPlus \cite{barsoum2016training}, SFEW 2.0 \cite{dhall2015video}, RAF-DB \cite{li2017reliable}, and AffectNet \cite{mollahosseini2017affectnet}. The first two datasets contain grayscale images, while the last three datasets contain RGB images. Specifically, CK+ \cite{lucey2010extended} includes 593 video sequences from 123 subjects. Following previous work \cite{jiang2022disentangling}, we extract the last three frames of each sequence to construct the dataset. RAF-DB \cite{li2017reliable} contains thousands of samples with basic and compound expressions, and this paper focuses on basic emotions. AffectNet \cite{mollahosseini2017affectnet} has 8 labels and each label contains 500 samples for evaluation. For FERPlus \cite{barsoum2016training} and SFEW 2.0 \cite{dhall2015video}, we use the official test set for performance evaluation.

\begin{table*}[t]
	\centering
	\renewcommand\arraystretch{1.26}
	\caption{Prompt template for different tasks. Please replace ``{Candidate labels}'' in the following prompts into candidate labels of each dataset. This paper uses batch-wise calling strategy to handle request limits. By default, we set the batch size to 20 for image-level inputs and 6 for video-level inputs.}
	\label{Table2}
	\scalebox{0.8}{
		\begin{tabular}{p{4.2cm}|p{17.6cm}}
			\hline
			Task & Prompt \\
			\hline
			
			\multirow{4}{*}{Visual Sentiment Analysis} & Please play the role of an emotion recognition expert. We provide 20 images. Please recognize sentiments evoked by these images (i.e., guess how viewer might emotionally feel after seeing these images.) If there is a person in the image, ignore that person's identity. For each image, please sort the provided categories from high to low according to the similarity with the image. Here are the optional categories: \{Candidate labels\}. The output format should be \{'name':, 'result':\} for each image. \\
			
			\hline
			
			\multirow{4}{*}{Tweet Sentiment Analysis} & Please play the role of an emotion recognition expert. We provide 20 image-text pairs. Please analyze how he will feel if he post this image-text pair on social media. If there is a person in the image, ignore that person's identity. For each image-text pair, please sort the provided categories from high to low according to the similarity with the input image-text pair. Here are the optional categories: \{Candidate labels\}. The output format should be \{'name':, 'result':\} for each image-text pair. \\
			
			\hline
			
			\multirow{4}{*}{Micro-expression Rec} & Please play the role of a micro-expression recognition expert. We provide 20 images. For each image, please sort the provided categories from high to low according to the similarity with the input image. The expression may not be obvious, please pay attention to the details of the face. Here are the optional categories: \{Candidate labels\}. Please ignore the speaker's identity and focus on the facial expression. The output format should be \{'name':, 'result':\} for each image. \\
			
			\hline
			
			\multirow{4}{*}{Facial Emotion Rec.} & Please play the role of a facial expression classification expert. We provide 20 images. For each image, please sort the provided categories from high to low according to the top 5 similarity with the input image. Here are the optional categories: \{Candidate labels\}. Please ignore the speaker's identity and focus on the facial expression. The output format should be \{'name':, 'result':\} for each image. \\
			
			\hline
			
			\multirow{4}{*}{Dynamic Facial Emotion Rec.} & Please play the role of a video expression classification expert. We provide 6 videos, each with 3 temporally uniformly sampled frames. For each video, please sort the provided categories from high to low according to the top 5 similarity with the input video. Here are the optional categories: \{Candidate labels\}. Please ignore the speaker's identity and focus on the facial expression. The output format should be \{'name':, 'result':\} for each video. \\
			
			\hline
			
			\multirow{5}{*}{Multimodal Emotion Rec.} & Please play the role of a video expression classification expert. We provide 6 videos, each with the speaker's content and 3 temporally uniformly sampled frames. Please ignore the speaker's identity and focus on their emotions. For each video, please sort the provided categories from high to low according to the top 5 similarity with the input video. Here are the optional categories: \{Candidate labels\}. Please ignore the speaker's identity and focus on their emotions. The output format should be \{'name':, 'result':\} for each video. \\
			
			\hline
		\end{tabular}
	}
\end{table*}

\textbf{Micro-expression Recognition} aims to identify subtle changes in human faces. Early works have proved the feasibility of recognizing micro-expressions from an apex frame \cite{li2020joint}. Therefore, we use the apex frame in our evaluation. Following previous works, we adopt the same label processing method and concentrate on major emotions. Specifically, CASME \cite{yan2013casme} comprises 195 samples across 8 categories, and we focus on the performance of four main labels: \emph{tense}, \emph{disgust}, \emph{repression}, and \emph{surprise}. CASME II \cite{yan2014casme} includes 247 samples collected from 26 subjects, and we focus on five major labels: \emph{happiness}, \emph{surprise}, \emph{disgust}, \emph{repression}, and \emph{others}. SAMM \cite{davison2016samm} consists of 159 samples, and our evaluation is limited to labels with more than 10 samples: \emph{anger}, \emph{contempt}, \emph{happiness}, \emph{surprise}, \emph{others}.

\textbf{Dynamic Facial Emotion Recognition} focuses on more challenging image sequences. In this task, we conduct experiments on four benchmark datasets: FERV39k \cite{wang2022ferv39k}, RAVDESS \cite{livingstone2018ryerson}, eNTERFACE05 \cite{martin2006enterface}, and DFEW \cite{jiang2020dfew}. For the first three datasets, we evaluate the performance on the official test set. DFEW \cite{jiang2020dfew} contains 5 folds with a total of 11,697 samples. To reduce the evaluation cost, we only report the results of fold 1 (fd1), consistent with previous works \cite{wang2022dpcnet, chen2023static}.

\textbf{Multimodal Emotion Recognition} focuses on three benchmark datasets: CH-SIMS \cite{zadeh2017tensor}, CMU-MOSI \cite{yu2020ch}, and MER-MULTI \cite{lian2023mer}. The first two datasets offer sentiment intensity scores. In this paper, we concentrate on the negative/positive classification task, where positive and negative classes are assigned for $< 0$ and $> 0$ scores, respectively. MER-MULTI is a subset of the MER2023 dataset \cite{lian2023mer}, providing both discrete and dimensional labels for each sample. In this paper, we focus on the recognition performance of discrete emotions.

\begin{algorithm}[t]
	\caption{GPT-4V Calling Strategy}
	\label{alg-1}
	\KwIn{Dataset $\mathcal{D}=\{x_i\}_{i=1}^{N}$ with ${N}$ samples, the maximum number of calls $T_{\max}$.}
	\KwOut{Prediction results for all samples.}
	\BlankLine
	
	\SetKwProg{Fn}{Def}{:}{}
	\SetKwFunction{FSuba}{Func}
	\SetKwFunction{FSubb}{Main}
	
	\Fn{\FSubb{$\mathcal{D}$}}{
		
		Split $\mathcal{D}$ into batches $\mathcal{D}=\{\mathcal{B}_i\}_{i=1}^{N_b}$;
		
		\For{$i = 1, \cdots, N_b$}{
			
			\emph{\textcolor[rgb]{0.93,0.0,0.47}{\# batch-wise calling}}
			
			\FSuba{$\mathcal{B}_i$}; 
			
		}
		
		\KwRet
		
	}
	
	~\\
	
	\Fn{\FSuba{$\mathcal{B}_i$}}{
		
		\emph{\textcolor[rgb]{0.93,0.0,0.47}{\# repeated calling}}
		
		\For{$t = 1, \cdots, T_{\max}$}{
			
			Call GPT-4V for $\mathcal{B}_i$ and get response $\mathcal{O}_i$;
			
			\If{meet requirements}{
				
				Store the response $\mathcal{O}_i$;
				
				\KwRet
			}
			
		}
		
		\emph{\textcolor[rgb]{0.93,0.0,0.47}{\# recursive calling}}
		
		\If{$|\mathcal{B}_i| \geq 2$}{
			
			Divide $\mathcal{B}_i$ into two equal parts: $\mathcal{C}_1$ and $\mathcal{C}_2$;
			
			\FSuba{$\mathcal{C}_1$};
			
			\FSuba{$\mathcal{C}_2$};
			
		}
		
		\KwRet
		
	}
	
\end{algorithm}

\section{GPT-4V Calling Strategy}
\label{sec4}
This paper evaluates the performance of the GPT-4V API, ``gpt-4-vision-preview''. GER tasks involve diverse modalities, such as images, text, video, and audio. However, GPT-4V only supports image and text inputs. To process video data, we sample the video and convert it into multiple images. To process audio data, we try converting the audio to a mel-spectrogram, which is a typical visual representation of the frequency content of an audio signal over time, with the frequency axis scaled according to the mel scale \cite{hermansky1990perceptual}. However, GPT-4V fails to generate proper responses for the mel-spectrogram. Therefore, the primary focus of this paper revolves around images, text, and video. In this section, we design a calling strategy specifically for GER tasks, containing batch-wise, repeated, and recursive calling. Pseudo-code is summarized in Algorithm \ref{alg-1}.

\textbf{Batch-wise Calling.} 
GPT-4V API has three limitations on requests: tokens per minute (TPM), requests per minute (RPM), and requests per day (RPD), which introduces additional requirements to our prompt design. To meet the constraints of RPM and RPD, we follow previous work \cite{wu2023gpt4vis} and adopt batch-wise inputs. That is, we feed multiple samples into GPT-4V and generate all results in one request. However, a large batch size may result in the total number of tokens exceeding the TPM limitation. Additionally, it increases task difficulty and may lead to incorrect responses. For example, if we input 100 samples in a request, GPT-4V might only generate predictions for 96 samples. Therefore, we set the batch size to 20 for image-level inputs and 6 for video-level inputs, aiming to satisfy TPM, RPM, and RPD limitations simultaneously. Prompts for different tasks can be found in Table \ref{Table2}.

\textbf{Repeated Calling.}
GER tasks often trigger security checks, causing GPT-4V to refuse to provide a response. This is attributed to the inclusion of visual sentiment analysis and human emotion recognition. The former task contains violent and bloody images. In the latter task, human identities are also considered as sensitive information. To reduce rejected cases, we require GPT-4V to ignore speaker identity (see Table \ref{Table2}), but it still triggers security errors. Interestingly, these errors sometimes occur randomly. For example, although all images are human-centric, some pass security checks while others fail. Alternatively, a sample may initially fail the check but pass upon retry. Therefore, we make multiple calls to the rejected batches. 

\textbf{Recursive Calling.}
During the evaluation, we notice that a batch-wise input may fail security checks, but splitting it into smaller parts can sometimes pass checks. Therefore, for the consistently failing batch, we split it into two smaller mini-batches and feed them into GPT-4V separately. This operation is repeated until further splitting is no longer feasible.

\textbf{Combination.} 
Our strategy combines batch-wise, repeated, and recursive calling. More details can be found in Algorithm \ref{alg-1}. A qualified response $\mathcal{O}$ should satisfy two conditions. First, it should not trigger the security check. Second, it should contain the correct number of prediction results. Figures \ref{Figure8}$\sim$\ref{Figure13} provide examples of qualified responses for different GER tasks.

\section{Results and Discussion}
\label{sec5}
This section presents and discusses results from three levels: dataset-wise, class-wise, and sample-wise. Specifically, we first present the results of heuristics baselines, supervised systems, and GPT-4V. Then, we reveal the temporal modeling and multimodal fusion capabilities of GPT-4V. Following that, we analyze the system's stability and conduct an in-depth class-wise performance analysis. Next, we assess the system's robustness against changes in prompt templates and color spaces. Finally, we visualize the rejected and erroneously predicted samples.

\begin{table}[t]
	\centering
	\renewcommand\tabcolsep{2pt}
	\caption{Performance comparison on visual sentiment analysis.}
	\label{Table3}
	\scalebox{0.8}{
		\begin{tabular}{l|cccc}
			\hline
			{Model} & {Twitter I (5/4/3)} & {Twitter II} & {Abstract} & {ArtPhoto} \\ 
			\hline
			SentiBank \cite{borth2013large} 				&71.32/68.28/66.63 &65.93 &64.95 &67.74 \\
			PAEF \cite{zhao2014exploring} 					&72.90/69.61/67.92 &77.51 &70.05 &67.85 \\
			DeepSentiBank \cite{chen2014deepsentibank}  	&76.35/70.15/71.25 &70.23 &71.19 &68.73 \\
			PCNN \cite{you2015robust}  						&82.54/76.52/76.36 &77.68 &70.84 &70.96 \\
			VGGNet \cite{simonyan2015very} 					&83.44/78.67/75.49 &71.79 &68.86 &67.61 \\
			AR (Concat) \cite{yang2018visual} 				&\textbf{88.65}/\textbf{85.10}/\textbf{81.06} &\textbf{80.48} &\textbf{76.03} &\textbf{74.80} \\
			\hline 
			Random  										&50.74/49.75/50.50 &50.37 & 50.88 & 50.41 \\
			Majority 										&66.82/62.51/61.33 &78.61 & 61.23 & 53.06 \\
			\rowcolor{lightgray}
			GPT-4V 											& \textbf{97.81}/\textbf{94.63}/\textbf{90.71} &\textbf{87.95} &\textbf{71.81} & \textbf{80.40} \\
			\hline
		\end{tabular}
	}
\end{table}

\begin{table}[t]
	\centering
	\renewcommand\tabcolsep{10.0pt}
	\caption{Performance comparison on micro-expression recognition.}
	\label{Table4}
	\scalebox{0.8}{
		\begin{tabular}{l|ccc}
			\hline
			Model	& CASME  & CASME II	& SAMM \\ 
			\hline
			LBP-SIP \cite{wang2015lbp} & 36.84  & 46.56 & 36.76 \\
			STRCN-A \cite{xia2019spatiotemporal} & 40.93  & 45.26 & 32.85 \\
			STRCN-G \cite{xia2019spatiotemporal} & 59.65  & 63.37 & 53.48 \\ 
			VGGMag \cite{li2018can}  & 60.23  & 63.21 & 36.00 \\
			LGCcon \cite{li2020joint}    & 60.82  & 65.02 & 40.90 \\
			TSCNN \cite{song2019recognizing}   & \textbf{73.88}  & \textbf{80.97} & \textbf{71.76} \\ 
			\hline
			Random   & 24.09 & 18.33 & 20.37 \\
			Majority & \textbf{40.34} & \textbf{40.17} & \textbf{42.22} \\
			\rowcolor{lightgray}
			GPT-4V & 36.93& 14.64 & 17.04 \\
			\hline
		\end{tabular}
	}
\end{table}

\begin{table}[t]
	\centering
	\caption{Performance comparison on tweet sentiment analysis.}
	\label{Table5}
	\scalebox{0.8}{
		\begin{tabular}{l|cc}
			\hline
			{Model} & {MVSA-Single} & {MVSA-Multiple} \\ 
			\hline
			CNN-Multi \cite{cai2015convolutional} 		& 61.20 & 66.39 \\
			DNN-LR \cite{yu2016visual} 					& 61.42 & 67.86 \\
			MultiSentiNet \cite{xu2017multisentinet} 	& 69.84 & 68.86 \\
			CoMN \cite{xu2018co} 						& 70.51 & 69.92 \\
			MVAN \cite{yang2020image} 					& 72.98 & 72.36 \\
			ITIN \cite{zhu2022multimodal} 				& \textbf{75.19} & \textbf{73.52} \\
			\hline 
			Random  									& 33.96 & 33.46 \\
			Majority 									& 55.68 & 66.65 \\
			\rowcolor{lightgray}
			GPT-4V 										& \textbf{61.25} & \textbf{66.82} \\
			\hline
		\end{tabular}
	}
\end{table}

\begin{table}[t]
	\centering
	\caption{Performance comparison on multimodal emotion recognition. These results come from MERBench \cite{lian2024merbench}, which reproduces different algorithms using the same feature set to ensure a fair comparison.}
	\label{Table6}
	\scalebox{0.8}{
		\begin{tabular}{l|ccc}
			\hline
			Model	& CMU-MOSI & CH-SIMS & MER-MULTI \\ 
			\hline
			MFM \cite{tsai2018learning}     & 78.34 & 87.14 & 70.48 \\
			MISA \cite{hazarika2020misa}    & 79.08 & 89.82 & 82.42 \\
			MFN \cite{zadeh2018memory}      & 77.78 & 87.00 & 77.40 \\
			MMIM \cite{han2021improving}    & 79.38 & 88.68 & 81.01 \\ 
			TFN \cite{zadeh2017tensor}      & 79.63 & 90.56 & 82.52 \\
			MulT  \cite{tsai2019multimodal} & \textbf{81.41} & \textbf{91.07} & \textbf{82.94} \\
			\hline 
			Random   &51.33 &51.23 & 17.87 \\
			Majority &42.37 &50.47 & 10.40 \\
			\rowcolor{lightgray}
			GPT-4V &\textbf{80.43} & \textbf{81.24} & \textbf{65.39}  \\
			\hline
		\end{tabular}
	}
\end{table}

\begin{table*}[t]
	\centering
	\caption{Performance comparison on facial emotion recognition.}
	\label{Table7}
	\scalebox{0.8}{
		\begin{tabular}{lc|lc|lc|lc|lc}
			\hline
			Model	& CK+ 	& Model & SFEW 2.0 & Model	& RAF-DB & Model	& FERPlus 	& Model & AffectNet \\
			\hline
			Cross-VAE \cite{wu2020cross} & 94.96 						& IACNN \cite{meng2017identity} &50.98 				&SCN \cite{wang2020suppressing} 	& 87.03				& RAN\cite{wang2020region} & 89.16					& SCN \cite{wang2020suppressing}  	&60.23 \\
			IA-gen \cite{yang2018identity} & 96.57  					& DLP-CNN \cite{li2017reliable} &51.05				&EfficientFace \cite{zhao2021robust}& 88.36				& SCN \cite{wang2020suppressing} & 89.39			& MA-Net \cite{zhao2021learning} 	&60.29 \\
			ADFL \cite{bai2019disentangled} & 98.17 					& TDTLN \cite{yan2019cross}	&53.10					&MA-Net \cite{zhao2021learning}  	& 88.42				& EAC \cite{zhang2022learn} & 89.64					& ARM \cite{shi2021learning}  		&61.33 \\
			IDFERM \cite{liu2019hard} & 98.35 							& DAN \cite{wen2023distract} &53.18					&TransFER \cite{xue2021transfer} 	& 90.91 			& KTN\cite{li2021adaptively} & 90.49				& DAN \cite{wen2023distract}  		&62.09 \\
			TER-GAN \cite{ali2021facial} & 98.47 						& RAN \cite{wang2020region} &54.19	 				&POSTER \cite{zheng2023poster}		& 92.05				& TransFER\cite{xue2021transfer} & 90.83			& POSTER \cite{zheng2023poster}	  	&63.34	\\
			IPD-FER \cite{jiang2022disentangling} & \textbf{98.65}		& FaceCaps \cite{wu2021facecaps} & \textbf{58.50}				&POSTER++ \cite{mao2023poster}		& \textbf{92.21} 	& POSTER \cite{zheng2023poster} & \textbf{91.62}	& POSTER++ \cite{mao2023poster}  	&\textbf{63.77} \\
			\hline
			Random & 14.61 & Random & 13.98 & Random & 14.57 & Random & 12.61 & Random & 12.73 \\
			Majority & 25.38 & Majority & 19.77 & Majority & 38.64 & Majority & 35.75 & Majority & 12.50\\
			\rowcolor{lightgray}
			GPT-4V & \textbf{69.72} & GPT-4V& \textbf{57.24} & GPT-4V & \textbf{75.81} & GPT-4V & \textbf{64.25} & GPT-4V & \textbf{42.77} \\ 
			\hline
		\end{tabular}
	}
\end{table*}

\begin{table*}[t]
	\centering
	\caption{Performance comparison on dynamic facial emotion recognition. Since GPT-4V does not support video input directly, we uniformly sample frames from the videos and sequentially feed these frames into GPT-4V. In this table, we test 2-frame (2-frm) and 3-frame (3-frm) sampling manners.}
	\label{Table8}
	\scalebox{0.8}{
		\begin{tabular}{lc|lc|lc|lc}
			\hline
			{Model} & {DFEW}  &{Model} & {FERV39k} & {Model} & {RAVDESS} & {Model} & {eNTERFACE05} \\ 
			\hline
			C3D \cite{tran2015learning}  		&56.37  			& M3DFEL \cite{wang2023rethinking} 		&47.67 			&AV-LSTM \cite{ghaleb2019multimodal} 	& 65.80 			& STA-FER \cite{pan2019deep}  			& 42.98 \\
			ResNet50-LSTM \cite{wang2022dpcnet} &63.32				& STT \cite{ma2022spatio} 				&48.11 			&MCBP \cite{su2020msaf} 				& 71.32 			& EC-LSTM \cite{miyoshi2021enhanced}  	& 49.26 \\
			DPCNet  \cite{wang2022dpcnet} 		&65.78 				& IAL \cite{li2023intensity}  			&48.54 			&MSAF \cite{su2020msaf} 				& 74.86 			& FAN \cite{meng2019frame}  			& 51.44 \\
			SVFAP \cite{sun2023svfap}  			&74.81 				& MAE-DFER \cite{sun2023mae} 			&52.07			&SVFAP \cite{sun2023svfap} 				& 75.01 			& Graph-Tran \cite{zhao2022spatial} 	& 54.62 \\
			MAE-DFER \cite{sun2023mae}  		&74.88 				& SVFAP \cite{sun2023svfap} 			&52.29 			&MAE-DFER \cite{sun2023mae} 			& 75.56 			& SVFAP \cite{sun2023svfap} 			& 60.54 \\
			S2D \cite{chen2023static}  			&\textbf{76.16} 	& S2D \cite{chen2023static} 			&\textbf{52.56} &CFN-SR \cite{fu2021cross}  			& \textbf{75.76} 	& MAE-DFER \cite{sun2023mae} 			& \textbf{61.64} \\
			\hline
			Random  							&14.43 & Random 							&14.41 &Random   							& 11.86 & Random   								& 16.87\\
			Majority 							&22.81 & Majority 							&24.79 &Majority 							& 13.33 & Majority   							& 16.71\\
			\rowcolor{lightgray}
			GPT-4V (2-frm) 						&{43.80} & GPT-4V (2-frm) 					&{34.29} &GPT-4V (2-frm) 					& 22.29 & GPT-4V (2-frm) 			& 23.78 \\
			\rowcolor{lightgray}
			GPT-4V (3-frm) 						&\textbf{54.80} & GPT-4V (3-frm) 					& \textbf{38.72} &GPT-4V (3-frm) 					& \textbf{34.31} & GPT-4V (3-frm) 					& \textbf{33.28} \\
			\hline
		\end{tabular}
	}
\end{table*}

\subsection{Main Results} 
Besides GPT-4V, we introduce two heuristic baselines: random guessing and majority guessing. For the former, we randomly select a label from candidate labels. For the latter, we choose the majority label as the prediction. For both baselines, we run experiments 10 times and report average results.

Table \ref{Table3} shows the results of visual sentiment analysis. We observe that GPT-4V outperforms supervised systems on most datasets. This superior performance can be attributed to its strong visual understanding ability, coupled with its reasoning capability, allowing GPT-4V to accurately infer the emotional states evoked by images. But for micro-expression recognition (see Table \ref{Table4}), GPT-4V exhibits poor performance, even worse than heuristic baselines. These results suggest that GPT-4V is designed for general-purpose domains (i.e., emotions that can be perceived by normal people). It is not well-suited for tasks that require professional knowledge like micro-expressions.

Tables \ref{Table5}$\sim$\ref{Table8} present the results of tweet sentiment analysis, multimodal emotion recognition, facial emotion recognition, and dynamic facial emotion recognition, respectively. To process video data, we uniformly sample frames and feed these frames into GPT-4V sequentially. To reduce call costs, we sample up to three frames per video. Despite a performance gap still existing between GPT-4V and supervised systems, it is worth noting that GPT-4V can significantly outperform heuristic baselines, demonstrating its potential in emotion recognition.

In Table \ref{Table6}, we observe a larger performance gap between supervised systems and GPT-4V on MER-MULTI compared to CMU-MOSI and CH-SIMS. This can be attributed to the fact that supervised systems incorporate acoustic information, while GPT-4V does not support audio input. Given that audio is more crucial in MER-MULTI \cite{lian2024merbench}, GPT-4V loses more information in this dataset, leading to limited performance.

\subsection{Temporal Modeling Ability}
To reduce GPT-4V call costs, this paper samples up to three frames per video. In this section, we further reveal the impact of sampling numbers. In Table \ref{Table8}, the performance improves when we increase the sampling number from two to three, indicating GPT-4V's temporal understanding ability. Additionally, it is worth noting that despite setting the sampling number to three, some key frames may still be ignored. Therefore, there is a possibility for further improvement by sampling more frames, and we leave this discussion for subsequent researchers.

\subsection{Multimodal Fusion Ability}
This section evaluates GPT-4V's multimodal fusion ability. Among all tasks, \emph{tweet emotion recognition} and \emph{multimodal emotion recognition} provide multimodal information. Hence, we conduct experiments on these tasks. Table \ref{Table9} shows unimodal and multimodal results. In general, multimodal results outperform unimodal results, showcasing GPT-4V's ability to integrate and exploit multi-modalities. But for CMU-MOSI, we notice a slight decrease in multimodal results compared to unimodal results. This can be attributed to CMU-MOSI mainly relying on lexical content to convey emotions \cite{lian2024merbench}, and the incorporation of visual clues may introduce interference.

\begin{table}[t]
	\centering
	\caption{Multimodal fusion ability. We report unimodal and multimodal results for tweet sentiment analysis (TSA) and multimodal emotion recognition (MER). Here, TSA takes text-image inputs and MER takes text-video inputs.}
	\label{Table9}
	\scalebox{0.8}{
		\begin{tabular}{cc|ccc}
			\hline
			Task & Dataset & Text & Image/Video & Fusion \\ 
			\hline
			
			\multirow{2}{*}{TSA} 
			&MVSA-Single   & 57.65 & 58.68 & \textbf{61.25} \\
			&MVSA-Multiple & 62.71 & 63.35 & \textbf{66.82} \\
			
			\hline
			
			\multirow{3}{*}{MER} 
			& CH-SIMS   &70.07  		&76.13  & \textbf{81.24} \\
			& CMU-MOSI  &\textbf{82.32} &51.17  & 80.43 		 \\
			& MER-MULTI &34.57  		&46.23  & \textbf{65.39} \\
			
			\hline
		\end{tabular}
	}
\end{table}

\subsection{System Stability}
\begin{figure}[t]
	\begin{center}
		\subfloat[identical predictions]{
			\label{Figure3-1}
			\centering
			\includegraphics[width=0.476\linewidth]{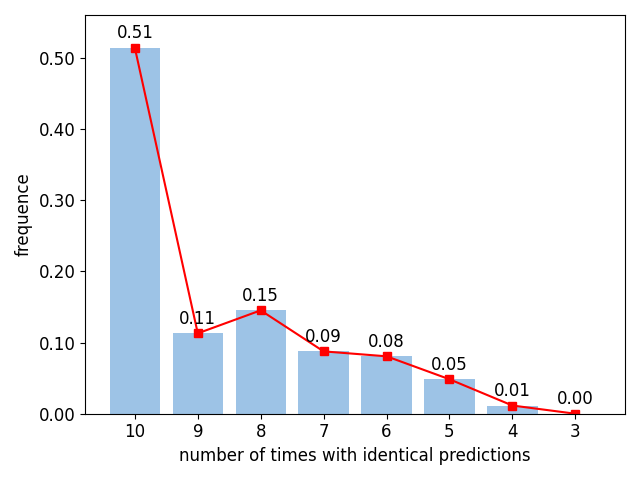}
		} 
		\subfloat[test accuracy]{
			\label{Figure3-2}
			\centering
			\includegraphics[width=0.476\linewidth]{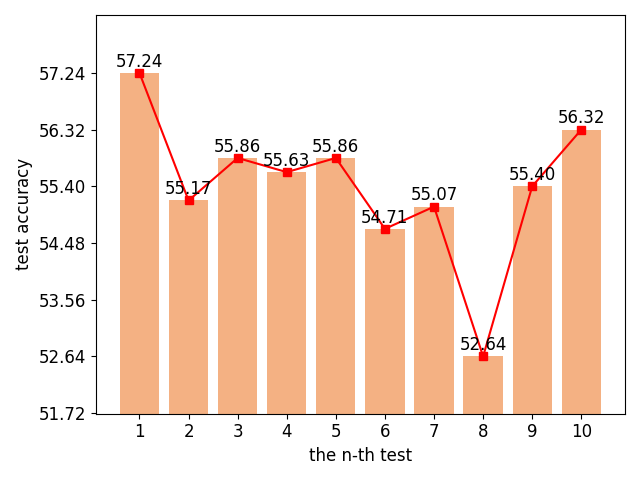}
		}
	\end{center}
	\caption{System stability test. We run GPT-4V 10 times, calculate the frequency with identical predictions, and report the test accuracy for different runs.}
	\label{Figure3}
\end{figure}

This section evaluates GPT-4V's prediction stability. During the experiments, we run GPT-4V 10 times for each sample in SFEW 2.0. Figure \ref{Figure3-1} shows the frequency of identical predictions. Specifically, we assume that GPT-4V predicts \emph{negative} 8 times and \emph{negative} 2 times. Therefore, it predicts the same label at most $c=8$ times. Then, we compute $c$ for all samples and calculate its frequency. In Figure \ref{Figure3-2}, we report the test accuracy for each run. We observe that although more than 50\% of the samples exhibit the same result across 10 runs, there are also some samples with different prediction results across distinct runs, leading to fluctuations in test accuracy. Meanwhile, we observe a 4.60\% performance gap between the best and worst runs. Therefore, GPT-4V exhibits certain instability. We recommend that subsequent researchers evaluate GPT-4V multiple times and use majority voting to get final results.

\subsection{Class-wise Performance Analysis}
\begin{figure*}[t]
	\begin{center}
		\subfloat[Twitter I(5)]{
			\label{Figure4-1}
			\centering
			\includegraphics[width=0.134\linewidth, trim=46 0 46 0]{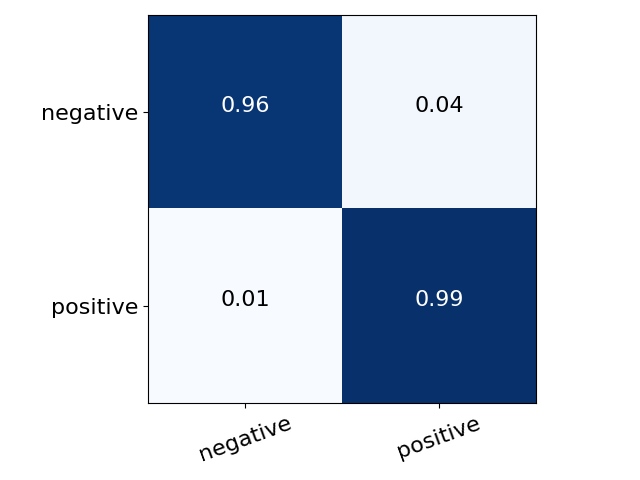}
		}
		\subfloat[Twitter II]{
			\label{Figure4-2}
			\centering
			\includegraphics[width=0.134\linewidth, trim=46 0 46 0]{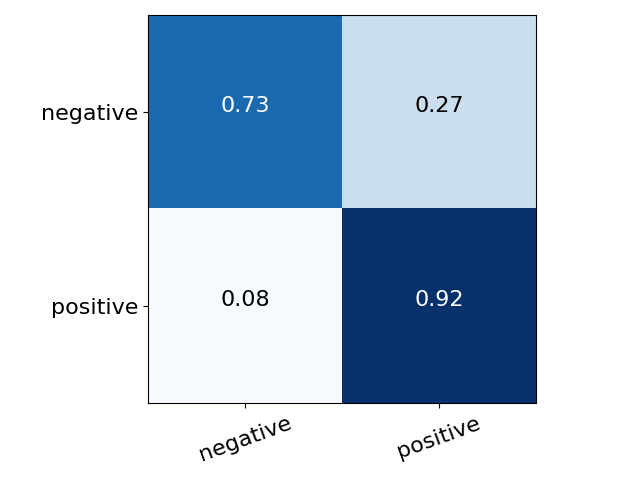}
		}
		\subfloat[Abstract]{
			\label{Figure4-3}
			\centering
			\includegraphics[width=0.134\linewidth, trim=46 0 46 0]{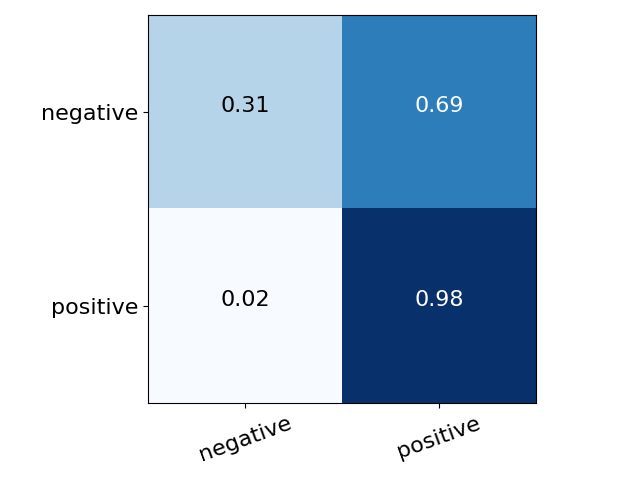}
		}
		\subfloat[ArtPhoto]{
			\label{Figure4-4}
			\centering
			\includegraphics[width=0.134\linewidth, trim=46 0 46 0]{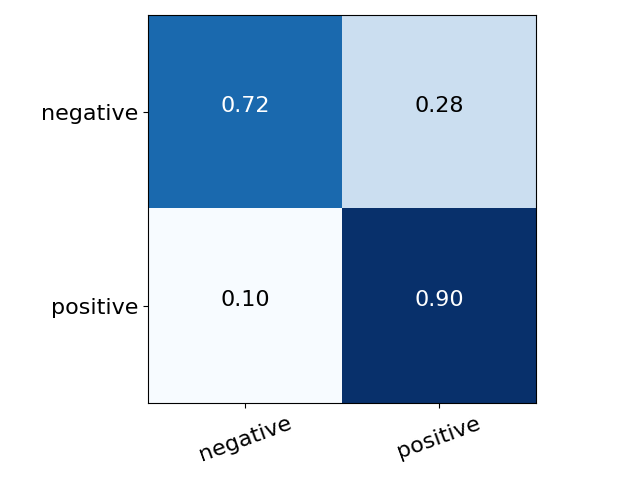}
		}
		\subfloat[MVSA-Single]{
			\label{Figure4-5}
			\centering
			\includegraphics[width=0.134\linewidth, trim=46 0 46 0]{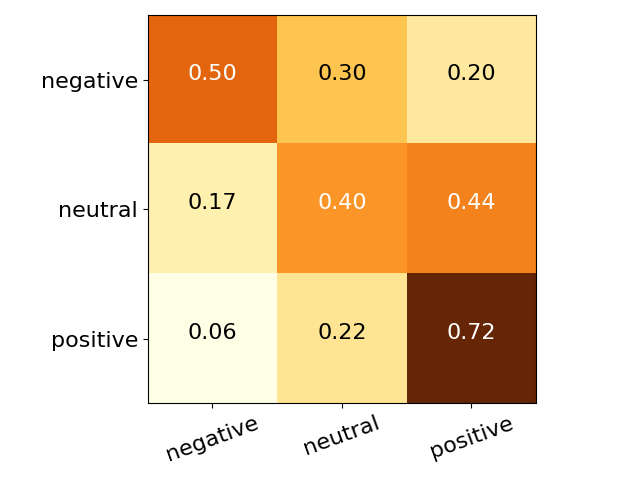}
		}
		\subfloat[MVSA-Multiple]{
			\label{Figure4-6}
			\centering
			\includegraphics[width=0.134\linewidth, trim=46 0 46 0]{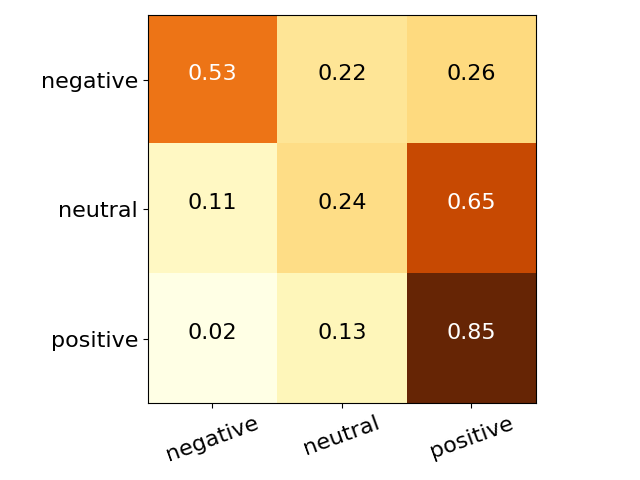}
		}
		\subfloat[CASME]{
			\label{Figure4-7}
			\centering
			\includegraphics[width=0.134\linewidth, trim=46 0 46 0]{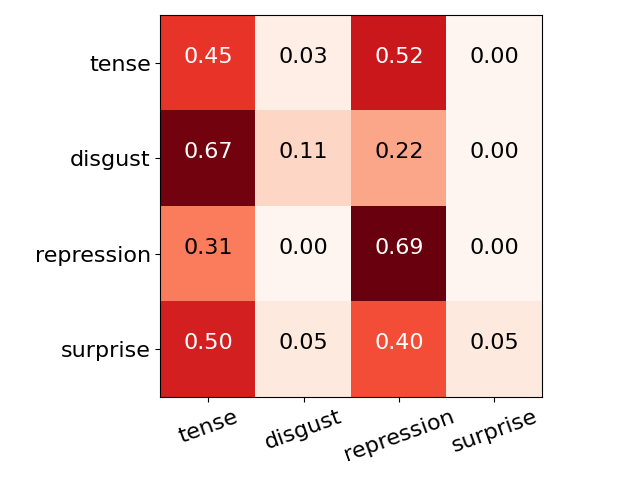}
		}
		
		\subfloat[CASME II]{
			\label{Figure4-8}
			\centering
			\includegraphics[width=0.134\linewidth, trim=46 0 46 0]{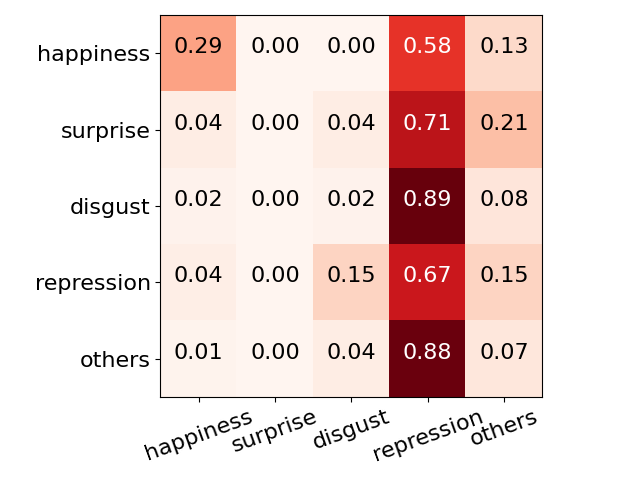}
		}
		\subfloat[SAMM]{
			\label{Figure4-9}
			\centering
			\includegraphics[width=0.134\linewidth, trim=46 0 46 0]{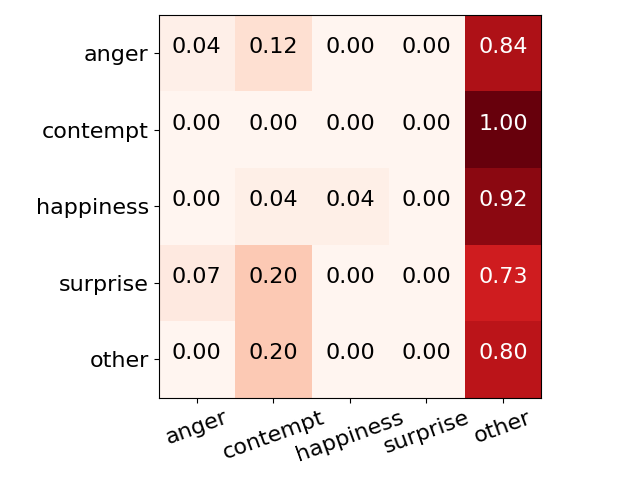}
		}
		\subfloat[CK+]{
			\label{Figure4-10}
			\centering
			\includegraphics[width=0.134\linewidth, trim=46 0 46 0]{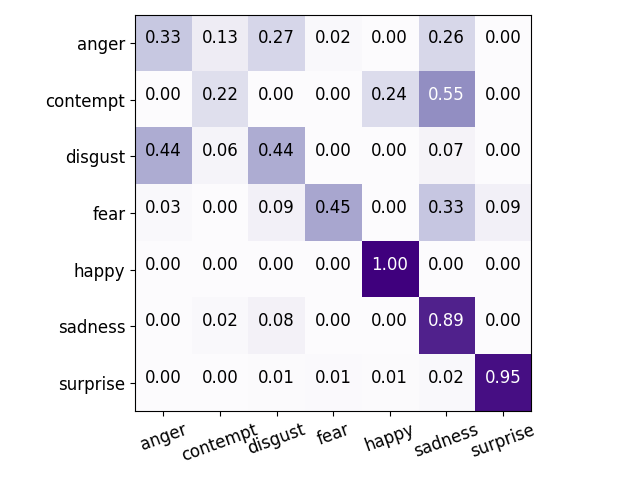}
		}
		\subfloat[SFEW 2.0]{
			\label{Figure4-11}
			\centering
			\includegraphics[width=0.134\linewidth, trim=46 0 46 0]{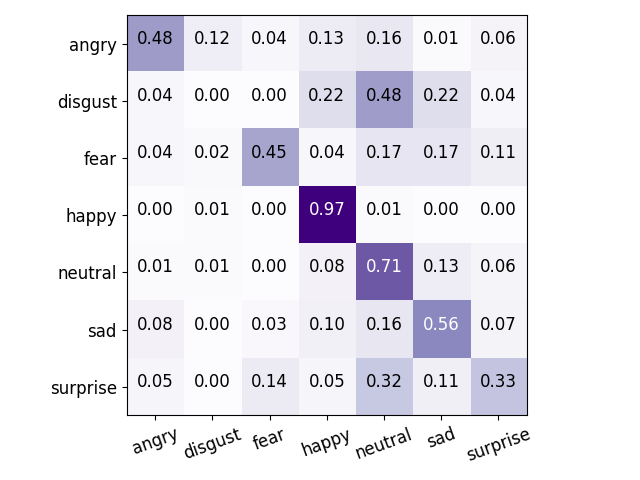}
		}
		\subfloat[FERPlus]{
			\label{Figure4-134}
			\centering
			\includegraphics[width=0.134\linewidth, trim=46 0 46 0]{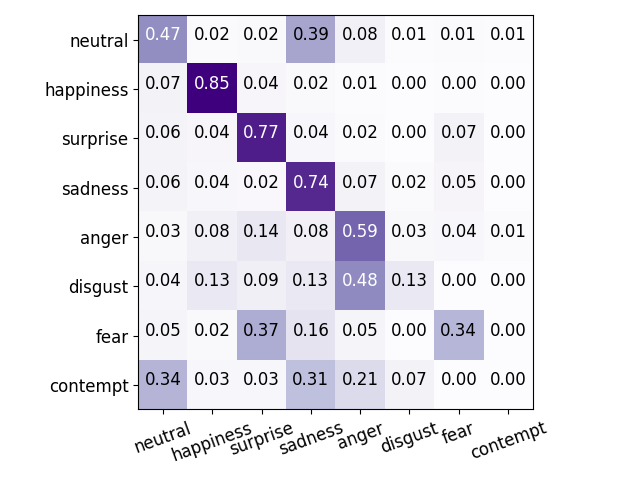}
		}
		\subfloat[RAF-DB]{
			\label{Figure4-12}
			\centering
			\includegraphics[width=0.134\linewidth, trim=46 0 46 0]{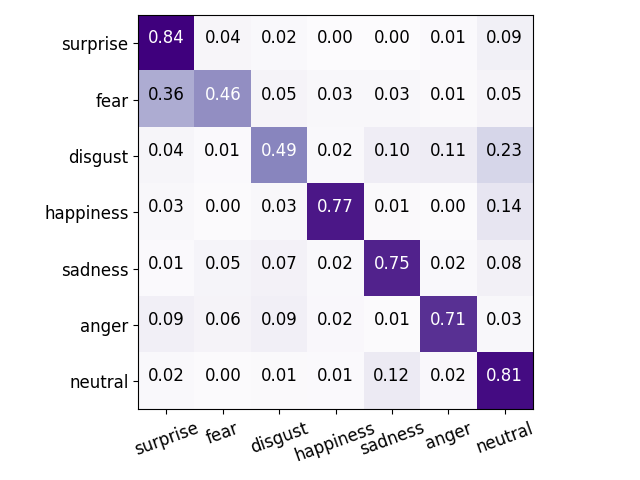}
		}
		\subfloat[AffectNet]{
			\label{Figure4-14}
			\centering
			\includegraphics[width=0.134\linewidth, trim=46 0 46 0]{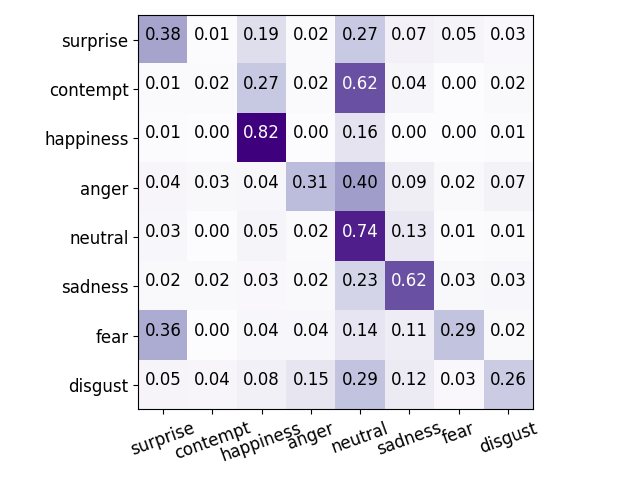}
		}
		
		\subfloat[DFEW]{
			\label{Figure4-15}
			\centering
			\includegraphics[width=0.134\linewidth, trim=46 0 46 0]{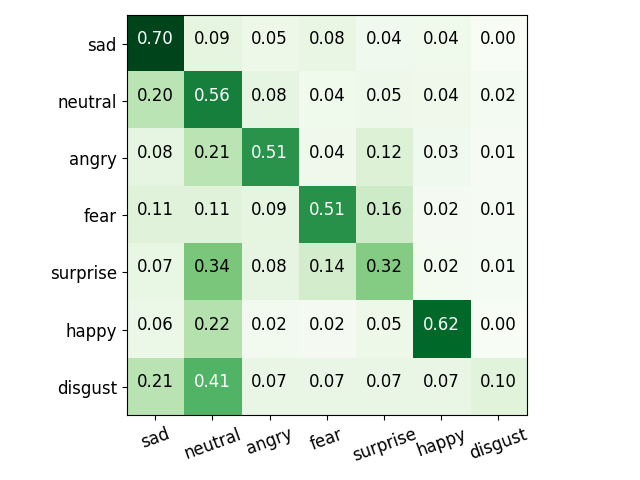}
		}
		\subfloat[FERV39k]{
			\label{Figure4-16}
			\centering
			\includegraphics[width=0.134\linewidth, trim=46 0 46 0]{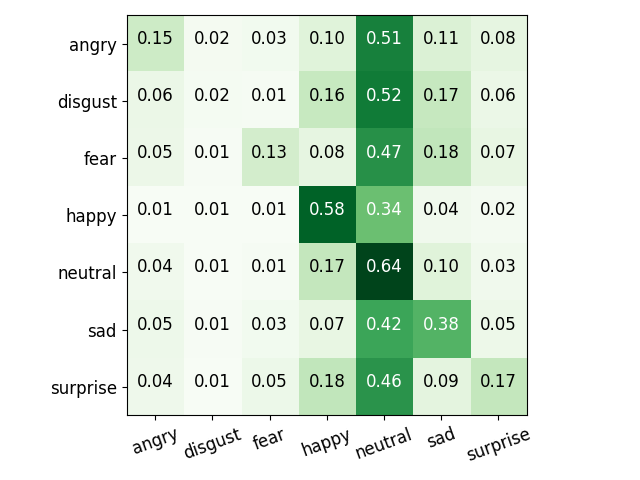}
		}
		\subfloat[RAVDESS]{
			\label{Figure4-17}
			\centering
			\includegraphics[width=0.134\linewidth, trim=46 0 46 0]{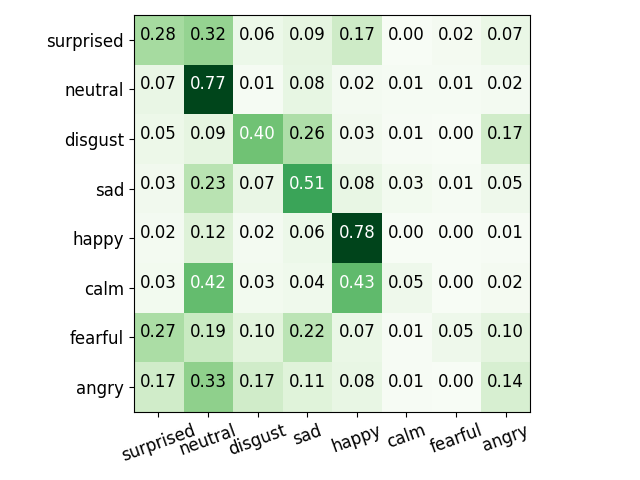}
		}
		\subfloat[eNTERFACE05]{
			\label{Figure4-18}
			\centering
			\includegraphics[width=0.134\linewidth, trim=46 0 46 0]{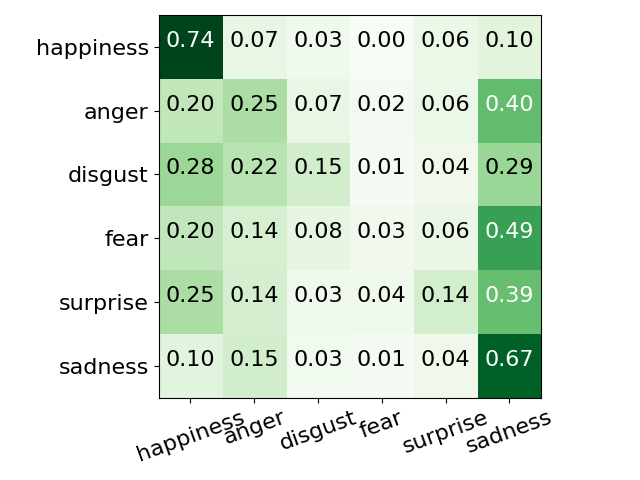}
		}
		\subfloat[CMU-MOSI]{
			\label{Figure4-19}
			\centering
			\includegraphics[width=0.134\linewidth, trim=46 0 46 0]{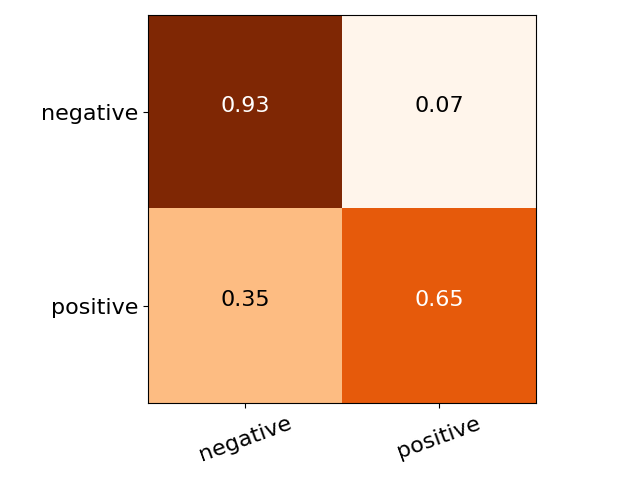}
		}
		\subfloat[CH-SIMS]{
			\label{Figure4-20}
			\centering
			\includegraphics[width=0.134\linewidth, trim=46 0 46 0]{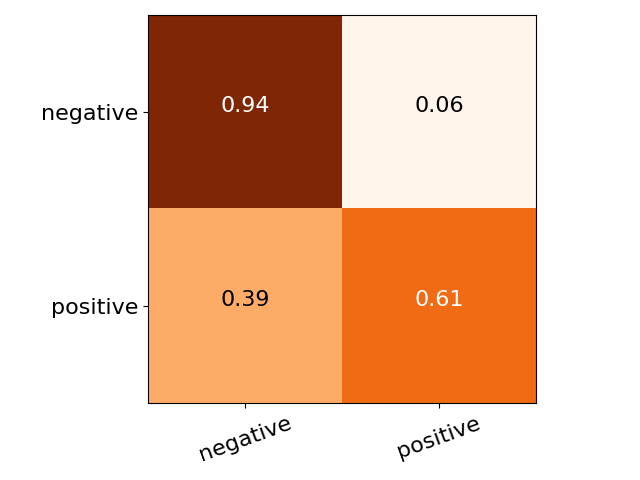}
		}
		\subfloat[MER-MULTI]{
			\label{Figure4-21}
			\centering
			\includegraphics[width=0.134\linewidth, trim=46 0 46 0]{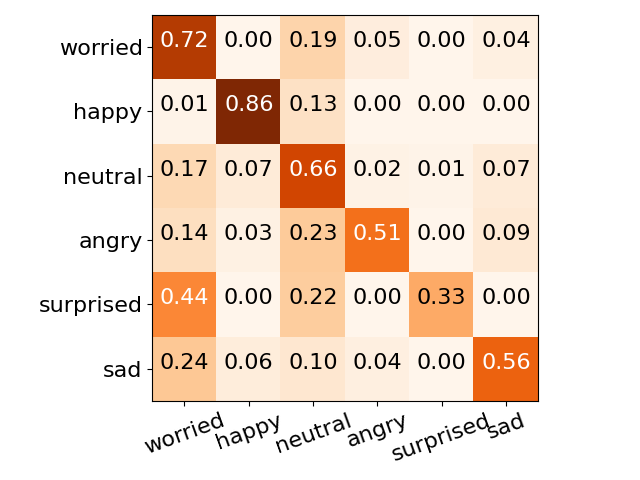}
		}
	\end{center}
	\caption{Confusion matrices for all evaluation datasets. Different colors represent distinct tasks.}
	\label{Figure4}
\end{figure*}

In Figure \ref{Figure4}, we visualize confusion matrices and perform a class-wise performance analysis. For visual sentiment analysis, GPT-4V generally performs well in \emph{positive} and \emph{negative} classes, showcasing its visual understanding ability. However, it exhibits lower results on Abstract compared to other datasets. This may be attributed to the fact that GPT-4V is mainly trained on natural images. The domain gap between abstract and natural images leads to increased difficulty and limited performance. For tweet sentiment analysis, GPT-4V performs relatively poorly in identifying \emph{neutral} sentiment and often misclassifies it as \emph{positive} or \emph{negative}. This may be due to GPT-4V's ability to detect subtle emotional cues in image-text pairs.

\begin{table*}[t]
	\centering
	\renewcommand\arraystretch{1}
	\caption{Robustness to prompt template changes. We conduct experiments on the facial emotion recognition task.}
	\label{Table10}
	\scalebox{0.8}{
		\begin{tabular}{p{0.6cm}|p{19cm}}
			\hline
			ID & Prompt Template \\
			\hline
			
			\multirow{3}{*}{\#1} & Please play the role of a facial expression classification expert. We provide 20 images. For each image, please sort the provided categories from high to low according to the top 5 similarity with the input image. Here are the optional categories: \{Candidate labels\}. Please ignore the speaker's identity and focus on the facial expression. The output format should be \{'name':, 'result':\} for each image. \\
			
			\hline
			
			\multirow{3}{*}{\#2} & We provide 20 images. For each image, please sort the provided categories from high to low according to the top 5 similarity with the input image. Here are the optional categories: \{Candidate labels\}. Please ignore the speaker's identity and focus on the facial expression. The output format should be \{'name':, 'result':\} for each image. \\
			
			\hline
			
			\multirow{3}{*}{\#3} & Please play the role of a facial expression classification expert. We provide 20 images. For each image, please select the most likely category according to the correlation with the input image. Here are the optional categories: \{Candidate labels\}. Please ignore the speaker's identity and focus on the facial expression. The output format should be \{'name':, 'result':\} for each image. \\
			
			\hline
		\end{tabular}
	}
\end{table*}

For micro-expression recognition, GPT-4V prefers \emph{tense} and \emph{repression}. Compared to other labels (e.g., \emph{surprise} and \emph{disgust}), \emph{tense} and \emph{repression} are less noticeable and closer to \emph{neutral}. These results indicate that GPT-4V tends to view micro-expressions as \emph{neutral}, causing it to perform poorly on this task. For facial expression recognition, GPT-4V tends to exhibit lower accuracy on \emph{contempt}, \emph{disgust}, and \emph{fear}, while achieving higher accuracy on \emph{happiness}. This phenomenon is related to the clarity of emotion definition. Since \emph{contempt}, \emph{disgust}, and \emph{fear} have less consensus between different annotations, they are inherently more difficult to recognize \cite{mollahosseini2017affectnet, picard2001toward}.

For dynamic facial emotion recognition, most samples tend to be predicted as \emph{neutral}, which is more obvious in FERV39. We attribute this phenomenon to our sampling operation. For each video, we sample up to three frames. GPT-4V tends to predict the emotion as \emph{neutral} if there is no emotion-related information within these frames. For multimodal emotion recognition, GPT-4V performs well on \emph{worry} and \emph{happiness}. For \emph{worry}, it has a strong reliance on lexical information. Differently, \emph{happiness} is mainly expressed through facial expressions \cite{lian2023mer}. To achieve good performance, GPT-4V should use different predominant modalities for distinct emotions, showcasing its multimodal fusion capabilities.

\subsection{Robustness to Template Change}
Previous works have pointed out that pre-trained language models (such as BERT and RoBERTa) are sensitive to prompt templates, with different templates yield varying results \cite{mao2022biases}. This raises the question: is GPT-4V also sensitive to template changes? In this section, we conduct experiments on SFEW 2.0 and compare the performance of different templates. In Table \ref{Table10}, Template\#1 is our default template. Compared with Template\#1, Template\#2 deletes ``Please play the role of a facial expression classification expert'' and Template\#3 changes the selection of the top 5 classes into the selection of the most likely class. Experimental results demonstrate that Template\#2 (57.93\%) and Template\#3 (54.25\%) can achieve similar performance compared to Template\#1 (57.24\%). Therefore, we can conclude that GPT-4V is robust to template changes.

\subsection{Robustness to Color Space}
\label{sec:color}
\begin{figure}[t]
	\centering
	\includegraphics[width=0.76\linewidth]{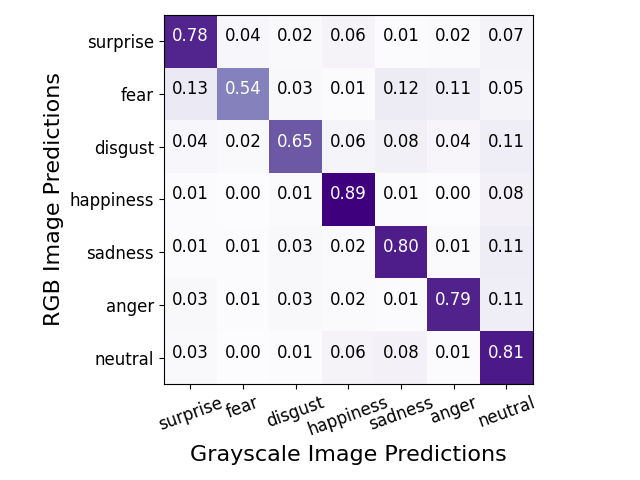}
	\caption{Prediction consistency on RGB and grayscale images.}
	\label{Figure5}
\end{figure}

This section evaluates the robustness of GPT-4V to color space changes. We conduct experiments on RAF-DB and convert RGB images into grayscale images. We observe that GPT-4V achieves nearly identical results on grayscale (74.28\%) and RGB images (75.81\%). These results demonstrate its robustness to color space changes. In Figure \ref{Figure5}, we further analyze the class-wise prediction consistency. We observe that the consistency rate is around 80\% for most classes except for \emph{fear} and \emph{disgust}. We further compare Figure \ref{Figure5} with the confusion matrix in Figure \ref{Figure4-12}. We observe similarities between these figures, where \emph{fear} and \emph{disgust} also perform poorly in emotion recognition. Therefore, we can conclude that poorly-performing categories in GPT-4V exhibit greater inconsistency in different color spaces. This may be attributed to poorly-performing categories with greater randomness across different runs.

\begin{figure}[t]
	\centering
	\includegraphics[width=\linewidth]{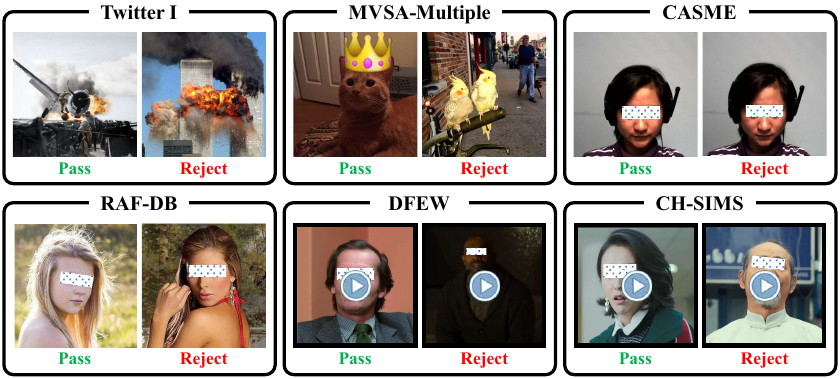}
	\caption{Passed and rejected examples by GPT-4V.}
	\label{Figure6}
\end{figure}

\subsection{Security Check}
\label{secuity-check}
GER includes visual sentiment analysis (containing images with violent and bloody content) and human emotion recognition (where human identity is considered sensitive information in GPT-4V). Therefore, it often triggers the security check. Figure \ref{Figure6} visualizes passed and rejected examples. In general, rejected examples contain humans or violent scenes. But there are some cases where examples containing such information can still pass checks. This observation suggests a certain degree of instability in the current security check used by GPT-4V.

\subsection{Case Study}
\begin{figure}[t]
	\centering
	\includegraphics[width=\linewidth]{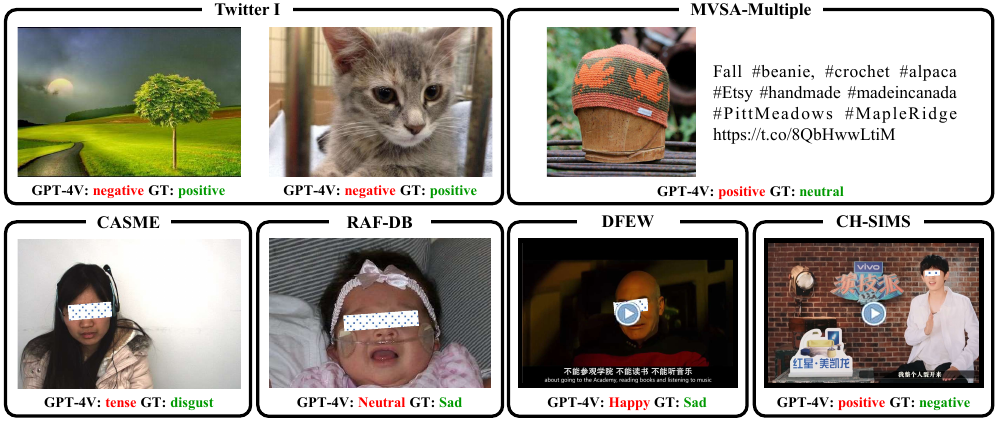}
	\caption{Error cases predicted by GPT-4V.}
	\label{Figure7}
\end{figure}

In Figure \ref{Figure7}, we visualize examples where GPT-4V makes incorrect predictions. More examples can be found in Figures \ref{Figure8}$\sim$\ref{Figure13} in the Appendix. For visual sentiment analysis (see Twitter I), we notice that GPT-4V can generate reasonable predictions. The main reason for these errors is the inherent multiple meanings of images. Take the image on the left as an example. One interpretation could emphasize that the sun rises, breaking through dark clouds and bathing the trees in sunlight, evoking positive emotions. Conversely, another explanation could emphasize that the sun is obstructed by dark clouds, triggering negative emotions. The same phenomenon can also be found in the picture on the right. On the one hand, there is a cute cat, but on the other hand, the cat is locked in a cage.

For tweet sentiment analysis (see MVSA-Multiple), the image shows a hat on wood, accompanied by descriptive text about the hat. Although the label is \emph{neutral}, GPT-4V predicts a \emph{positive} sentiment. We speculate that GPT-4V may interpret this image as depicting a comfortable environment and a nice hat, thereby eliciting a positive response.

For micro-expression recognition (see CASME), we observe that her emotion is not obvious. Without specialized knowledge, it is difficult to identify her emotion as \emph{disgust} and more likely to be predicted as \emph{neutral}. For CASME, there are four candidate labels: \emph{tense}, \emph{disgust}, \emph{repression}, and \emph{surprise}. Among them, \emph{tense} is closer to \emph{neutral}. Consequently, GPT-4V incorrectly predicts her emotion as \emph{tense}. In the future, we recommend that researchers provide clearer definitions of micro-expressions in the prompt. For example, offer typical clues to distinguish these micro-expressions. We believe this strategy can enhance the performance of GPT-4V on this task.

For facial emotion recognition (see RAF-DB), the image depicts a baby with an open mouth, indicating that she is crying. Meanwhile, she is wearing an oxygen mask, indicating that she is sick. Hence, her emotion is likely to be \emph{sad}. But GPT-4V incorrectly predicts her emotion as \emph{neutral}. This may be due to its inability to recognize the oxygen mask and infer her health status, indicating its limitations in understanding the details.

For dynamic facial emotion recognition (see DFEW), GPT-4V might predict the emotion as \emph{happiness} based on the raised corners of the mouth. But when considering the subtitle, ``cannot go to the academy, read books, and listen to music'', we can infer that he is more likely to be \emph{sad} due to these restrictions. Hence, the error made by GPT-4V could be attributed to the limitations of this task, as it relies only on the visual modality.

For multimodal emotion recognition (see CH-SIMS), GPT-4V predicts the emotion as \emph{positive} based on his facial expression. However, when taking into account the subtitle, ``I feel like I'm falling apart'', the content of his complaint shifts his emotion towards \emph{negative}. However, GPT-4V is unable to capture such emotional transitions, indicating its limitations in multimodal fusion under complex scenarios.

In summary, GPT-4V can provide relatively reasonable responses. These errors are mainly caused by the inherent multiple meanings of images or the limited capability to understand the details. Meanwhile, micro-expressions are different from regular emotions and GPT-4V faces challenges in recognizing them. Furthermore, its multimodal fusion ability may face challenges when different modalities express distinct emotions.

\section{Limitations and Future Directions}
\label{sec6}
In this section, we summarize the main challenges and limitations of GPT-4V as revealed by the evaluation results. Additionally, we discuss potential directions for future research.

\textbf{Performance.} In GER tasks, GPT-4V outperforms heuristic baselines but still lags behind supervised systems. This may be attributed to the vague definition of emotion \cite{lian2023explainable}. In the future, we plan to use few-shot prompts to enhance task clarity and GPT-4V's emotion understanding ability.

\textbf{Support Modality.} GPT-4V does not support audio. However, audio can convey emotions through prosodic information such as pitch, duration, and intensity. To better recognize emotions, GPT-4V should support audio in the future. Alternatively, we can use outside models to convert audio into lexical descriptions and use them as additional input for GPT-4V.

\textbf{Domain Gap.} GPT-4V is designed for general-purpose domains and performs poorly on micro-expressions that require specialized knowledge. For future work, we recommend that subsequent researchers can attempt to provide clearer definitions of micro-expressions in the prompt design.

\textbf{System Stability.} GPT-4V exhibits certain instability in prediction results, i.e., it may predict different labels for a sample. Therefore, GPT-4V should improve its system stability in the future. Alternatively, subsequent works should evaluate GPT-4V multiple times and use majority voting to get final results.

\textbf{Security Check Stability.} GPT-4V demonstrates certain instability in security checks. Specifically, a sample may initially fail the check but pass after a retry. To solve this problem, we design a more practical calling strategy in Section \ref{sec4}. In the future, GPT-4V should enhance its security check stability.

Despite the above limitations, GPT-4V has demonstrated promising results in emotion recognition and holds application potential in practical scenarios such as social media analysis, online education, and customer service. Different from previous solutions that required task-specific models tailored to each scenario, GPT-4V can serve as a universal model capable of minimizing application costs.

\section{Conclusion}
\label{sec7}
This paper conducts a comprehensive evaluation of GPT-4V in GER tasks. Our evaluation results demonstrate that GPT-4V exhibits strong visual understanding capabilities and achieves promising results in general-purpose emotion recognition. However, it performs poorly in micro-expression recognition which requires specialized knowledge. This paper also showcases GPT-4V's temporal modeling and multimodal fusion capabilities, along with its robustness to color space and prompt template changes. Furthermore, we evaluate its prediction consistency and security check stability and visualize error cases to reveal its limitations in emotion understanding. This paper can serve as a zero-shot benchmark and provide guidance for subsequent research on emotion recognition and MLLMs.

\section*{Acknowledgements}
This work is supported by the National Natural Science Foundation of China (NSFC) (No.62201572, No.61831022, No.62276259, No.62322120, No.U21B2010) and Beijing Municipal Science \& Technology Commission, Administrative Commission of Zhongguancun Science Park (No.Z211100004821013).

\bibliographystyle{elsarticle-num} 
\bibliography{mybib}

\begin{thebibliography}{100}
\expandafter\ifx\csname url\endcsname\relax
  \def\url#1{\texttt{#1}}\fi
\expandafter\ifx\csname urlprefix\endcsname\relax\def\urlprefix{URL }\fi
\expandafter\ifx\csname href\endcsname\relax
  \def\href#1#2{#2} \def\path#1{#1}\fi

\bibitem{you2017visual}
Q.~You, H.~Jin, J.~Luo, Visual sentiment analysis by attending on local image
  regions, in: Proceedings of the Thirty-First AAAI Conference on Artificial
  Intelligence, 2017, pp. 231--237.

\bibitem{lian2022smin}
Z.~Lian, B.~Liu, J.~Tao, Smin: Semi-supervised multi-modal interaction network
  for conversational emotion recognition, IEEE Transactions on Affective
  Computing (2022).

\bibitem{yang2023emoset}
J.~Yang, Q.~Huang, T.~Ding, D.~Lischinski, D.~Cohen-Or, H.~Huang, Emoset: A
  large-scale visual emotion dataset with rich attributes, in: Proceedings of
  the IEEE/CVF International Conference on Computer Vision, 2023, pp.
  20383--20394.

\bibitem{yang2023dawn}
Z.~Yang, L.~Li, K.~Lin, J.~Wang, C.-C. Lin, Z.~Liu, L.~Wang, The dawn of lmms:
  Preliminary explorations with gpt-4v (ision), arXiv preprint arXiv:2309.17421
  9 (2023) 1.

\bibitem{ortis2020survey}
A.~Ortis, G.~M. Farinella, S.~Battiato, Survey on visual sentiment analysis,
  IET Image Processing 14~(8) (2020) 1440--1456.

\bibitem{niu2016sentiment}
T.~Niu, S.~Zhu, L.~Pang, A.~El~Saddik, Sentiment analysis on multi-view social
  data, in: MultiMedia Modeling: 22nd International Conference, MMM 2016,
  Miami, FL, USA, January 4-6, 2016, Proceedings, Part II 22, Springer, 2016,
  pp. 15--27.

\bibitem{li2022deep}
Y.~Li, J.~Wei, Y.~Liu, J.~Kauttonen, G.~Zhao, Deep learning for
  micro-expression recognition: A survey, IEEE Transactions on Affective
  Computing (2022).

\bibitem{li2020deep}
S.~Li, W.~Deng, Deep facial expression recognition: A survey, IEEE transactions
  on affective computing 13~(3) (2020) 1195--1215.

\bibitem{wang2022ferv39k}
Y.~Wang, Y.~Sun, Y.~Huang, Z.~Liu, S.~Gao, W.~Zhang, W.~Ge, W.~Zhang, Ferv39k:
  A large-scale multi-scene dataset for facial expression recognition in
  videos, in: Proceedings of the IEEE/CVF Conference on Computer Vision and
  Pattern Recognition, 2022, pp. 20922--20931.

\bibitem{lian2024merbench}
Z.~Lian, L.~Sun, Y.~Ren, H.~Gu, H.~Sun, L.~Chen, B.~Liu, J.~Tao, Merbench: A
  unified evaluation benchmark for multimodal emotion recognition, arXiv
  preprint arXiv:2401.03429 (2024).

\bibitem{ekman1978facial}
P.~Ekman, W.~V. Friesen, Facial action coding system, Environmental Psychology
  \& Nonverbal Behavior (1978).

\bibitem{ekman1994strong}
P.~EKMAN, Strong evidence for universals in facial expressions: a reply to
  russell's mistaken critique, Psychological bulletin 115~(2) (1994) 268--287.

\bibitem{sebe2005multimodal}
N.~Sebe, I.~Cohen, T.~Gevers, T.~S. Huang, Multimodal approaches for emotion
  recognition: a survey, in: Internet Imaging VI, Vol. 5670, SPIE, 2005, pp.
  56--67.

\bibitem{lian2021ctnet}
Z.~Lian, B.~Liu, J.~Tao, Ctnet: Conversational transformer network for emotion
  recognition, IEEE/ACM Transactions on Audio, Speech, and Language Processing
  29 (2021) 985--1000.

\bibitem{wu2014survey}
C.-H. Wu, J.-C. Lin, W.-L. Wei, Survey on audiovisual emotion recognition:
  databases, features, and data fusion strategies, APSIPA Transactions on
  Signal and Information Processing 3 (2014).

\bibitem{you2015robust}
Q.~You, J.~Luo, H.~Jin, J.~Yang, Robust image sentiment analysis using
  progressively trained and domain transferred deep networks, in: Proceedings
  of the Twenty-Ninth AAAI Conference on Artificial Intelligence, 2015, pp.
  381--388.

\bibitem{borth2013large}
D.~Borth, R.~Ji, T.~Chen, T.~Breuel, S.-F. Chang, Large-scale visual sentiment
  ontology and detectors using adjective noun pairs, in: Proceedings of the
  21st ACM International Conference on Multimedia, 2013, pp. 223--232.

\bibitem{you2016building}
Q.~You, J.~Luo, H.~Jin, J.~Yang, Building a large scale dataset for image
  emotion recognition: the fine print and the benchmark, in: Proceedings of the
  Thirtieth AAAI Conference on Artificial Intelligence, 2016, pp. 308--314.

\bibitem{yan2013casme}
W.-J. Yan, Q.~Wu, Y.-J. Liu, S.-J. Wang, X.~Fu, Casme database: A dataset of
  spontaneous micro-expressions collected from neutralized faces, in:
  Proceedings of the 10th IEEE International Conference and Workshops on
  Automatic Face and Gesture Recognition (FG), IEEE, 2013, pp. 1--7.

\bibitem{yan2014casme}
W.-J. Yan, X.~Li, S.-J. Wang, G.~Zhao, Y.-J. Liu, Y.-H. Chen, X.~Fu, Casme ii:
  An improved spontaneous micro-expression database and the baseline
  evaluation, PLoS ONE 9~(1) (2014) e86041--e86041.

\bibitem{davison2016samm}
A.~K. Davison, C.~Lansley, N.~Costen, K.~Tan, M.~H. Yap, Samm: A spontaneous
  micro-facial movement dataset, IEEE Transactions on Affective Computing 9~(1)
  (2016) 116--129.

\bibitem{lucey2010extended}
P.~Lucey, J.~F. Cohn, T.~Kanade, J.~Saragih, Z.~Ambadar, I.~Matthews, The
  extended cohn-kanade dataset (ck+): A complete dataset for action unit and
  emotion-specified expression, in: IEEE Computer Society Conference on
  Computer Vision and Pattern Recognition Workshops, IEEE, 2010, pp. 94--101.

\bibitem{dhall2015video}
A.~Dhall, O.~Ramana~Murthy, R.~Goecke, J.~Joshi, T.~Gedeon, Video and image
  based emotion recognition challenges in the wild: Emotiw 2015, in:
  Proceedings of the 2015 ACM on International Conference on Multimodal
  Interaction, 2015, pp. 423--426.

\bibitem{li2017reliable}
S.~Li, W.~Deng, J.~Du, Reliable crowdsourcing and deep locality-preserving
  learning for expression recognition in the wild, in: Proceedings of the IEEE
  Conference on Computer Vision and Pattern Recognition, 2017, pp. 2852--2861.

\bibitem{barsoum2016training}
E.~Barsoum, C.~Zhang, C.~C. Ferrer, Z.~Zhang, Training deep networks for facial
  expression recognition with crowd-sourced label distribution, in: Proceedings
  of the 18th ACM International Conference on Multimodal Interaction, 2016, pp.
  279--283.

\bibitem{mollahosseini2017affectnet}
A.~Mollahosseini, B.~Hasani, M.~H. Mahoor, Affectnet: A database for facial
  expression, valence, and arousal computing in the wild, IEEE Transactions on
  Affective Computing 10~(1) (2017) 18--31.

\bibitem{jiang2020dfew}
X.~Jiang, Y.~Zong, W.~Zheng, C.~Tang, W.~Xia, C.~Lu, J.~Liu, Dfew: A
  large-scale database for recognizing dynamic facial expressions in the wild,
  in: Proceedings of the 28th ACM international conference on multimedia, 2020,
  pp. 2881--2889.

\bibitem{livingstone2018ryerson}
S.~R. Livingstone, F.~A. Russo, The ryerson audio-visual database of emotional
  speech and song (ravdess): A dynamic, multimodal set of facial and vocal
  expressions in north american english, PloS One 13~(5) (2018) e0196391.

\bibitem{martin2006enterface}
O.~Martin, I.~Kotsia, B.~Macq, I.~Pitas, The enterface'05 audio-visual emotion
  database, in: Proceedings of the 22nd International Conference on Data
  Engineering Workshops, IEEE, 2006, pp. 8--8.

\bibitem{zadeh2017tensor}
A.~Zadeh, M.~Chen, S.~Poria, E.~Cambria, L.-P. Morency, Tensor fusion network
  for multimodal sentiment analysis, in: Proceedings of the Conference on
  Empirical Methods in Natural Language Processing, 2017, pp. 1103--1114.

\bibitem{yu2020ch}
W.~Yu, H.~Xu, F.~Meng, Y.~Zhu, Y.~Ma, J.~Wu, J.~Zou, K.~Yang, Ch-sims: A
  chinese multimodal sentiment analysis dataset with fine-grained annotation of
  modality, in: Proceedings of the 58th Annual Meeting of the Association for
  Computational Linguistics, 2020, pp. 3718--3727.

\bibitem{lian2023mer}
Z.~Lian, H.~Sun, L.~Sun, K.~Chen, M.~Xu, K.~Wang, K.~Xu, Y.~He, Y.~Li, J.~Zhao,
  et~al., Mer 2023: Multi-label learning, modality robustness, and
  semi-supervised learning, in: Proceedings of the 31st ACM International
  Conference on Multimedia, 2023, pp. 9610--9614.

\bibitem{zhu2023minigpt}
D.~Zhu, J.~Chen, X.~Shen, X.~Li, M.~Elhoseiny, Minigpt-4: Enhancing
  vision-language understanding with advanced large language models, arXiv
  preprint arXiv:2304.10592 (2023).

\bibitem{liu2023visual}
H.~Liu, C.~Li, Q.~Wu, Y.~J. Lee, Visual instruction tuning, arXiv preprint
  arXiv:2304.08485 (2023).

\bibitem{li2023videochat}
K.~Li, Y.~He, Y.~Wang, Y.~Li, W.~Wang, P.~Luo, Y.~Wang, L.~Wang, Y.~Qiao,
  Videochat: Chat-centric video understanding, arXiv preprint arXiv:2305.06355
  (2023).

\bibitem{zhang2023speechgpt}
D.~Zhang, S.~Li, X.~Zhang, J.~Zhan, P.~Wang, Y.~Zhou, X.~Qiu, Speechgpt:
  Empowering large language models with intrinsic cross-modal conversational
  abilities, arXiv preprint arXiv:2305.11000 (2023).

\bibitem{su2023pandagpt}
Y.~Su, T.~Lan, H.~Li, J.~Xu, Y.~Wang, D.~Cai, Pandagpt: One model to
  instruction-follow them all, arXiv preprint arXiv:2305.16355 (2023).

\bibitem{alpaca_eval}
X.~Li, T.~Zhang, Y.~Dubois, R.~Taori, I.~Gulrajani, C.~Guestrin, P.~Liang,
  T.~B. Hashimoto, Alpacaeval: An automatic evaluator of instruction-following
  models, \url{https://github.com/tatsu-lab/alpaca_eval} (2023).

\bibitem{lu2024mathvista}
P.~Lu, H.~Bansal, T.~Xia, J.~Liu, C.~Li, H.~Hajishirzi, H.~Cheng, K.-W. Chang,
  M.~Galley, J.~Gao, Mathvista: Evaluating mathematical reasoning of foundation
  models in visual contexts, in: Proceedings of the International Conference on
  Learning Representations, {ICLR}, 2024, pp. 1--116.

\bibitem{lin2023mm}
K.~Lin, F.~Ahmed, L.~Li, C.-C. Lin, E.~Azarnasab, Z.~Yang, J.~Wang, L.~Liang,
  Z.~Liu, Y.~Lu, et~al., Mm-vid: Advancing video understanding with gpt-4v
  (ision), arXiv preprint arXiv:2310.19773 (2023).

\bibitem{wu2023can}
C.~Wu, J.~Lei, Q.~Zheng, W.~Zhao, W.~Lin, X.~Zhang, X.~Zhou, Z.~Zhao, Y.~Zhang,
  Y.~Wang, et~al., Can gpt-4v (ision) serve medical applications? case studies
  on gpt-4v for multimodal medical diagnosis, arXiv preprint arXiv:2310.09909
  (2023).

\bibitem{wu2023gpt4vis}
W.~Wu, H.~Yao, M.~Zhang, Y.~Song, W.~Ouyang, J.~Wang, Gpt4vis: What can gpt-4
  do for zero-shot visual recognition?, arXiv preprint arXiv:2311.15732 (2023).

\bibitem{xu2017multisentinet}
N.~Xu, W.~Mao, Multisentinet: A deep semantic network for multimodal sentiment
  analysis, in: Proceedings of the 2017 ACM on Conference on Information and
  Knowledge Management, 2017, pp. 2399--2402.

\bibitem{zhu2022multimodal}
T.~Zhu, L.~Li, J.~Yang, S.~Zhao, H.~Liu, J.~Qian, Multimodal sentiment analysis
  with image-text interaction network, IEEE Transactions on Multimedia (2022).

\bibitem{jiang2022disentangling}
J.~Jiang, W.~Deng, Disentangling identity and pose for facial expression
  recognition, IEEE Transactions on Affective Computing 13~(4) (2022)
  1868--1878.

\bibitem{li2020joint}
Y.~Li, X.~Huang, G.~Zhao, Joint local and global information learning with
  single apex frame detection for micro-expression recognition, IEEE
  Transactions on Image Processing 30 (2020) 249--263.

\bibitem{wang2022dpcnet}
Y.~Wang, Y.~Sun, W.~Song, S.~Gao, Y.~Huang, Z.~Chen, W.~Ge, W.~Zhang, Dpcnet:
  Dual path multi-excitation collaborative network for facial expression
  representation learning in videos, in: Proceedings of the 30th ACM
  International Conference on Multimedia, 2022, pp. 101--110.

\bibitem{chen2023static}
Y.~Chen, J.~Li, S.~Shan, M.~Wang, R.~Hong, From static to dynamic: Adapting
  landmark-aware image models for facial expression recognition in videos,
  arXiv preprint arXiv:2312.05447 (2023).

\bibitem{hermansky1990perceptual}
H.~Hermansky, Perceptual linear predictive (plp) analysis of speech, the
  Journal of the Acoustical Society of America 87~(4) (1990) 1738--1752.

\bibitem{zhao2014exploring}
S.~Zhao, Y.~Gao, X.~Jiang, H.~Yao, T.-S. Chua, X.~Sun, Exploring
  principles-of-art features for image emotion recognition, in: Proceedings of
  the 22nd ACM international conference on Multimedia, 2014, pp. 47--56.

\bibitem{chen2014deepsentibank}
T.~Chen, D.~Borth, T.~Darrell, S.-F. Chang, Deepsentibank: Visual sentiment
  concept classification with deep convolutional neural networks, arXiv
  preprint arXiv:1410.8586 (2014).

\bibitem{simonyan2015very}
K.~Simonyan, A.~Zisserman, Very deep convolutional networks for large-scale
  image recognition, in: Proceedings of the International Conference on
  Learning Representations, {ICLR}, 2015, pp. 1--14.

\bibitem{yang2018visual}
J.~Yang, D.~She, M.~Sun, M.-M. Cheng, P.~L. Rosin, L.~Wang, Visual sentiment
  prediction based on automatic discovery of affective regions, IEEE
  Transactions on Multimedia 20~(9) (2018) 2513--2525.

\bibitem{wang2015lbp}
Y.~Wang, J.~See, R.~C.-W. Phan, Y.-H. Oh, Lbp with six intersection points:
  Reducing redundant information in lbp-top for micro-expression recognition,
  in: Computer Vision--ACCV 2014: 12th Asian Conference on Computer Vision,
  Singapore, Singapore, November 1-5, 2014, Revised Selected Papers, Part I 12,
  Springer, 2015, pp. 525--537.

\bibitem{xia2019spatiotemporal}
Z.~Xia, X.~Hong, X.~Gao, X.~Feng, G.~Zhao, Spatiotemporal recurrent
  convolutional networks for recognizing spontaneous micro-expressions, IEEE
  Transactions on Multimedia 22~(3) (2019) 626--640.

\bibitem{li2018can}
Y.~Li, X.~Huang, G.~Zhao, Can micro-expression be recognized based on single
  apex frame?, in: 2018 25th IEEE International Conference on Image Processing
  (ICIP), IEEE, 2018, pp. 3094--3098.

\bibitem{song2019recognizing}
B.~Song, K.~Li, Y.~Zong, J.~Zhu, W.~Zheng, J.~Shi, L.~Zhao, Recognizing
  spontaneous micro-expression using a three-stream convolutional neural
  network, IEEE Access 7 (2019) 184537--184551.

\bibitem{cai2015convolutional}
G.~Cai, B.~Xia, Convolutional neural networks for multimedia sentiment
  analysis, in: Natural Language Processing and Chinese Computing: 4th CCF
  Conference, NLPCC 2015, Nanchang, China, October 9-13, 2015, Proceedings 4,
  Springer, 2015, pp. 159--167.

\bibitem{yu2016visual}
Y.~Yu, H.~Lin, J.~Meng, Z.~Zhao, Visual and textual sentiment analysis of a
  microblog using deep convolutional neural networks, Algorithms 9~(41) (2016)
  1--11.

\bibitem{xu2018co}
N.~Xu, W.~Mao, G.~Chen, A co-memory network for multimodal sentiment analysis,
  in: The 41st international ACM SIGIR conference on research \& development in
  information retrieval, 2018, pp. 929--932.

\bibitem{yang2020image}
X.~Yang, S.~Feng, D.~Wang, Y.~Zhang, Image-text multimodal emotion
  classification via multi-view attentional network, IEEE Transactions on
  Multimedia 23 (2020) 4014--4026.

\bibitem{tsai2018learning}
Y.-H.~H. Tsai, P.~P. Liang, A.~Zadeh, L.-P. Morency, R.~Salakhutdinov, Learning
  factorized multimodal representations, in: Proceedings of the 7th
  International Conference on Learning Representations, 2019, pp. 1--20.

\bibitem{hazarika2020misa}
D.~Hazarika, R.~Zimmermann, S.~Poria, Misa: Modality-invariant and-specific
  representations for multimodal sentiment analysis, in: Proceedings of the
  28th {ACM} International Conference on Multimedia, 2020, pp. 1122--1131.

\bibitem{zadeh2018memory}
A.~Zadeh, P.~P. Liang, N.~Mazumder, S.~Poria, E.~Cambria, L.-P. Morency, Memory
  fusion network for multi-view sequential learning, in: Proceedings of the
  {AAAI} Conference on Artificial Intelligence, 2018, pp. 5634--5641.

\bibitem{han2021improving}
W.~Han, H.~Chen, S.~Poria, Improving multimodal fusion with hierarchical mutual
  information maximization for multimodal sentiment analysis, in: Proceedings
  of the 2021 Conference on Empirical Methods in Natural Language Processing,
  2021, pp. 9180--9192.

\bibitem{tsai2019multimodal}
Y.-H.~H. Tsai, S.~Bai, P.~P. Liang, J.~Z. Kolter, L.-P. Morency,
  R.~Salakhutdinov, Multimodal transformer for unaligned multimodal language
  sequences, in: Proceedings of the 57th Conference of the Association for
  Computational Linguistics, 2019, pp. 6558--6569.

\bibitem{wu2020cross}
H.~Wu, J.~Jia, L.~Xie, G.~Qi, Y.~Shi, Q.~Tian, Cross-vae: Towards disentangling
  expression from identity for human faces, in: ICASSP 2020-2020 IEEE
  International Conference on Acoustics, Speech and Signal Processing (ICASSP),
  IEEE, 2020, pp. 4087--4091.

\bibitem{meng2017identity}
Z.~Meng, P.~Liu, J.~Cai, S.~Han, Y.~Tong, Identity-aware convolutional neural
  network for facial expression recognition, in: 2017 12th IEEE International
  Conference on Automatic Face \& Gesture Recognition (FG 2017), IEEE, 2017,
  pp. 558--565.

\bibitem{wang2020suppressing}
K.~Wang, X.~Peng, J.~Yang, S.~Lu, Y.~Qiao, Suppressing uncertainties for
  large-scale facial expression recognition, in: Proceedings of the IEEE/CVF
  conference on computer vision and pattern recognition, 2020, pp. 6897--6906.

\bibitem{wang2020region}
K.~Wang, X.~Peng, J.~Yang, D.~Meng, Y.~Qiao, Region attention networks for pose
  and occlusion robust facial expression recognition, IEEE Transactions on
  Image Processing 29 (2020) 4057--4069.

\bibitem{yang2018identity}
H.~Yang, Z.~Zhang, L.~Yin, Identity-adaptive facial expression recognition
  through expression regeneration using conditional generative adversarial
  networks, in: 2018 13th IEEE International Conference on Automatic Face \&
  Gesture Recognition (FG 2018), IEEE, 2018, pp. 294--301.

\bibitem{zhao2021robust}
Z.~Zhao, Q.~Liu, F.~Zhou, Robust lightweight facial expression recognition
  network with label distribution training, in: Proceedings of the AAAI
  Conference on Artificial Intelligence, 2021, pp. 3510--3519.

\bibitem{zhao2021learning}
Z.~Zhao, Q.~Liu, S.~Wang, Learning deep global multi-scale and local attention
  features for facial expression recognition in the wild, IEEE Transactions on
  Image Processing 30 (2021) 6544--6556.

\bibitem{bai2019disentangled}
M.~Bai, W.~Xie, L.~Shen, Disentangled feature based adversarial learning for
  facial expression recognition, in: 2019 IEEE International Conference on
  Image Processing (ICIP), IEEE, 2019, pp. 31--35.

\bibitem{yan2019cross}
K.~Yan, W.~Zheng, T.~Zhang, Y.~Zong, C.~Tang, C.~Lu, Z.~Cui, Cross-domain
  facial expression recognition based on transductive deep transfer learning,
  IEEE Access 7 (2019) 108906--108915.

\bibitem{zhang2022learn}
Y.~Zhang, C.~Wang, X.~Ling, W.~Deng, Learn from all: Erasing attention
  consistency for noisy label facial expression recognition, in: European
  Conference on Computer Vision, Springer, 2022, pp. 418--434.

\bibitem{shi2021learning}
J.~Shi, S.~Zhu, Z.~Liang, Learning to amend facial expression representation
  via de-albino and affinity, arXiv preprint arXiv:2103.10189 (2021).

\bibitem{liu2019hard}
X.~Liu, B.~V. Kumar, P.~Jia, J.~You, Hard negative generation for
  identity-disentangled facial expression recognition, Pattern Recognition 88
  (2019) 1--12.

\bibitem{wen2023distract}
Z.~Wen, W.~Lin, T.~Wang, G.~Xu, Distract your attention: Multi-head cross
  attention network for facial expression recognition, Biomimetics 8~(2) (2023)
  199.

\bibitem{xue2021transfer}
F.~Xue, Q.~Wang, G.~Guo, Transfer: Learning relation-aware facial expression
  representations with transformers, in: Proceedings of the IEEE/CVF
  International Conference on Computer Vision, 2021, pp. 3601--3610.

\bibitem{li2021adaptively}
H.~Li, N.~Wang, X.~Ding, X.~Yang, X.~Gao, Adaptively learning facial expression
  representation via cf labels and distillation, IEEE Transactions on Image
  Processing 30 (2021) 2016--2028.

\bibitem{ali2021facial}
K.~Ali, C.~E. Hughes, Facial expression recognition by using a disentangled
  identity-invariant expression representation, in: 2020 25th International
  Conference on Pattern Recognition (ICPR), IEEE, 2021, pp. 9460--9467.

\bibitem{zheng2023poster}
C.~Zheng, M.~Mendieta, C.~Chen, Poster: A pyramid cross-fusion transformer
  network for facial expression recognition, in: Proceedings of the IEEE/CVF
  International Conference on Computer Vision, 2023, pp. 3146--3155.

\bibitem{wu2021facecaps}
F.~Wu, C.~Pang, B.~Zhang, Facecaps for facial expression recognition, Computer
  Animation and Virtual Worlds 32~(3-4) (2021) 1--6.

\bibitem{mao2023poster}
J.~Mao, R.~Xu, X.~Yin, Y.~Chang, B.~Nie, A.~Huang, Poster v2: A simpler and
  stronger facial expression recognition network, arXiv preprint
  arXiv:2301.12149 (2023).

\bibitem{tran2015learning}
D.~Tran, L.~Bourdev, R.~Fergus, L.~Torresani, M.~Paluri, Learning
  spatiotemporal features with 3d convolutional networks, in: Proceedings of
  the IEEE international conference on computer vision, 2015, pp. 4489--4497.

\bibitem{wang2023rethinking}
H.~Wang, B.~Li, S.~Wu, S.~Shen, F.~Liu, S.~Ding, A.~Zhou, Rethinking the
  learning paradigm for dynamic facial expression recognition, in: Proceedings
  of the IEEE/CVF Conference on Computer Vision and Pattern Recognition, 2023,
  pp. 17958--17968.

\bibitem{ghaleb2019multimodal}
E.~Ghaleb, M.~Popa, S.~Asteriadis, Multimodal and temporal perception of
  audio-visual cues for emotion recognition, in: 2019 8th International
  Conference on Affective Computing and Intelligent Interaction (ACII), IEEE,
  2019, pp. 552--558.

\bibitem{pan2019deep}
X.~Pan, G.~Ying, G.~Chen, H.~Li, W.~Li, A deep spatial and temporal aggregation
  framework for video-based facial expression recognition, IEEE Access 7 (2019)
  48807--48815.

\bibitem{ma2022spatio}
F.~Ma, B.~Sun, S.~Li, Spatio-temporal transformer for dynamic facial expression
  recognition in the wild, arXiv preprint arXiv:2205.04749 (2022).

\bibitem{su2020msaf}
L.~Su, C.~Hu, G.~Li, D.~Cao, Msaf: Multimodal split attention fusion, arXiv
  preprint arXiv:2012.07175 (2020).

\bibitem{miyoshi2021enhanced}
R.~Miyoshi, N.~Nagata, M.~Hashimoto, Enhanced convolutional lstm with spatial
  and temporal skip connections and temporal gates for facial expression
  recognition from video, Neural Computing and Applications 33 (2021)
  7381--7392.

\bibitem{li2023intensity}
H.~Li, H.~Niu, Z.~Zhu, F.~Zhao, Intensity-aware loss for dynamic facial
  expression recognition in the wild, in: Proceedings of the AAAI Conference on
  Artificial Intelligence, 2023, pp. 67--75.

\bibitem{meng2019frame}
D.~Meng, X.~Peng, K.~Wang, Y.~Qiao, Frame attention networks for facial
  expression recognition in videos, in: 2019 IEEE international conference on
  image processing (ICIP), IEEE, 2019, pp. 3866--3870.

\bibitem{sun2023svfap}
L.~Sun, Z.~Lian, K.~Wang, Y.~He, M.~Xu, H.~Sun, B.~Liu, J.~Tao, Svfap:
  Self-supervised video facial affect perceiver, arXiv preprint
  arXiv:2401.00416 (2023).

\bibitem{sun2023mae}
L.~Sun, Z.~Lian, B.~Liu, J.~Tao, Mae-dfer: Efficient masked autoencoder for
  self-supervised dynamic facial expression recognition, in: Proceedings of the
  31st ACM International Conference on Multimedia, 2023, pp. 6110--6121.

\bibitem{zhao2022spatial}
R.~Zhao, T.~Liu, Z.~Huang, D.~P. Lun, K.-M. Lam, Spatial-temporal graphs plus
  transformers for geometry-guided facial expression recognition, IEEE
  Transactions on Affective Computing (2022).

\bibitem{fu2021cross}
Z.~Fu, F.~Liu, H.~Wang, J.~Qi, X.~Fu, A.~Zhou, Z.~Li, A cross-modal fusion
  network based on self-attention and residual structure for multimodal emotion
  recognition, arXiv preprint arXiv:2111.02172 (2021).

\bibitem{picard2001toward}
R.~W. Picard, E.~Vyzas, J.~Healey, Toward machine emotional intelligence:
  Analysis of affective physiological state, IEEE Transactions on Pattern
  Analysis and Machine Intelligence 23~(10) (2001) 1175--1191.

\bibitem{mao2022biases}
R.~Mao, Q.~Liu, K.~He, W.~Li, E.~Cambria, The biases of pre-trained language
  models: An empirical study on prompt-based sentiment analysis and emotion
  detection, IEEE Transactions on Affective Computing (2022).

\bibitem{lian2023explainable}
Z.~Lian, L.~Sun, M.~Xu, H.~Sun, K.~Xu, Z.~Wen, S.~Chen, B.~Liu, J.~Tao,
  Explainable multimodal emotion reasoning, arXiv preprint arXiv:2306.15401
  (2023).

\end{thebibliography}

\appendix

\begin{figure*}[t]
	\centering
	\includegraphics[width=0.8 \linewidth]{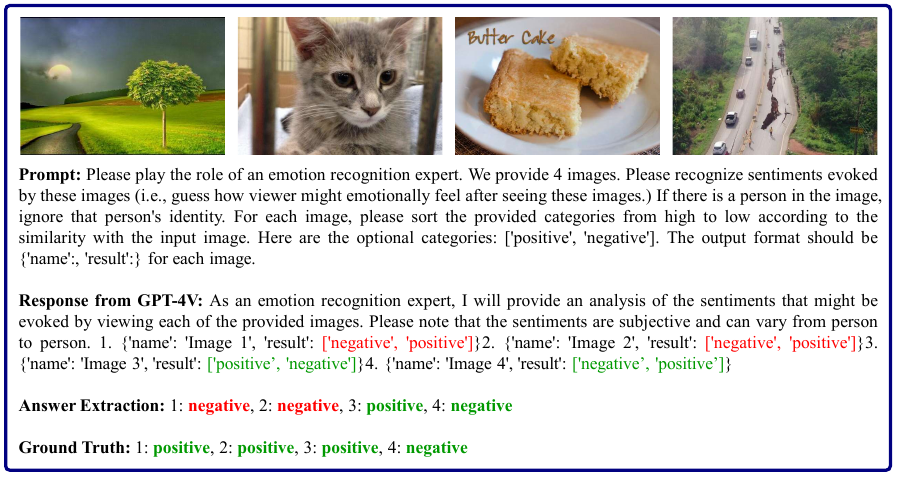}
	\caption{Case study on visual sentiment analysis. We show examples with \textcolor[rgb]{0,0.6,0}{correct} or \textcolor[rgb]{1,0,0}{wrong} answers generated by GPT-4V.}
	\label{Figure8}
\end{figure*}

\begin{figure*}[t]
	\centering
	\includegraphics[width=0.8 \linewidth]{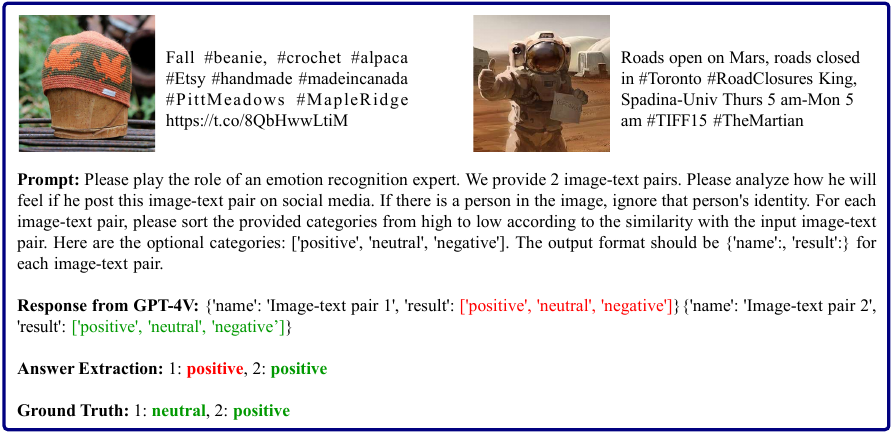}
	\caption{Case study on tweet sentiment analysis. We show examples with \textcolor[rgb]{0,0.6,0}{correct} or \textcolor[rgb]{1,0,0}{wrong} answers generated by GPT-4V.}
	\label{Figure9}
\end{figure*}

\begin{figure*}[t]
	\centering
	\includegraphics[width=0.8 \linewidth]{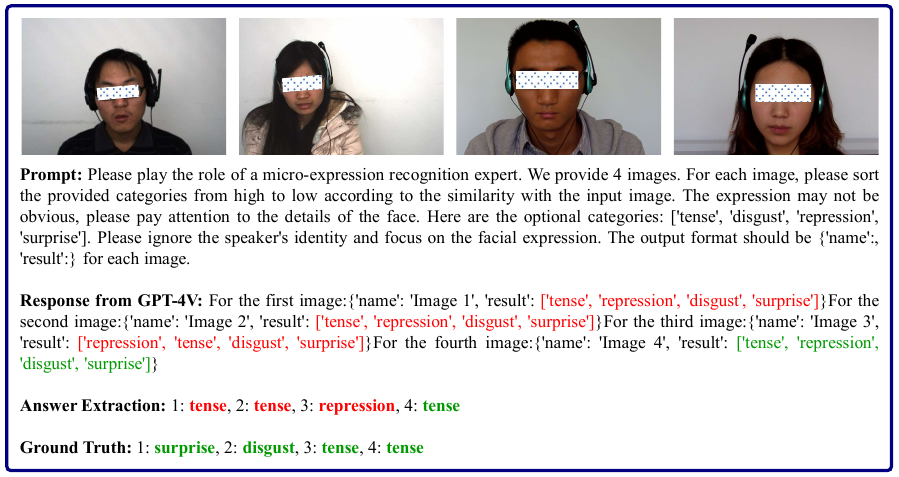}
	\caption{Case study on micro-expression recognition. We show examples with \textcolor[rgb]{0,0.6,0}{correct} or \textcolor[rgb]{1,0,0}{wrong} answers generated by GPT-4V.}
	\label{Figure10}
\end{figure*}

\begin{figure*}[t]
	\centering
	\includegraphics[width=0.8 \linewidth]{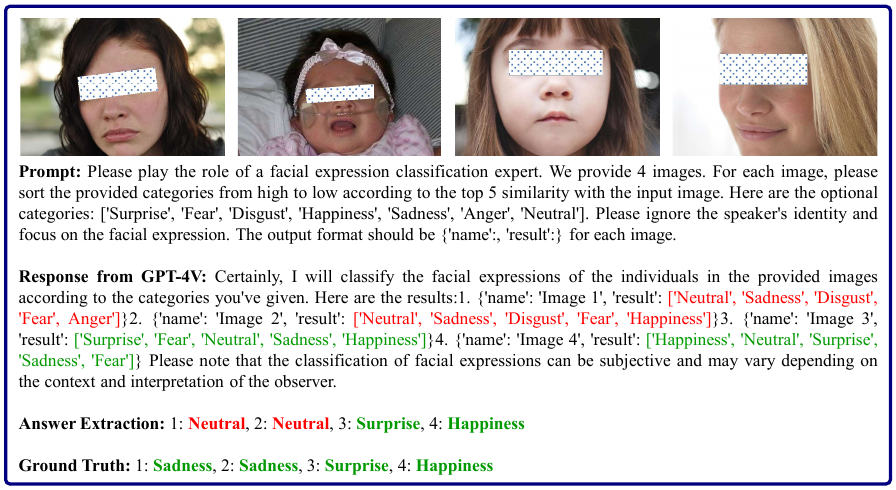}
	\caption{Case study for facial emotion recognition. We provide examples with \textcolor[rgb]{0,0.6,0}{correct} and \textcolor[rgb]{1,0,0}{wrong} answers generated by GPT-4V.}
	\label{Figure11}
\end{figure*}

\begin{figure*}[t]
	\centering
	\includegraphics[width=0.8 \linewidth]{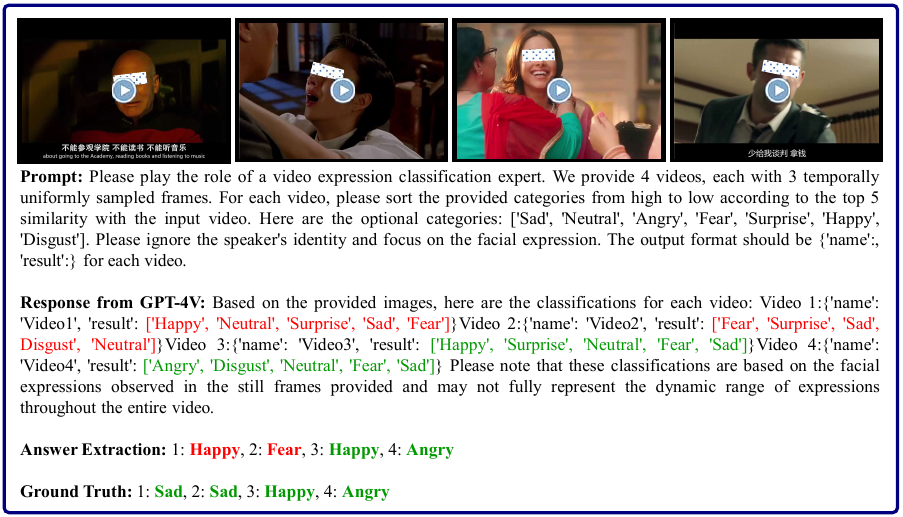}
	\caption{Case study for dynamic facial emotion recognition. We provide examples with \textcolor[rgb]{0,0.6,0}{correct} and \textcolor[rgb]{1,0,0}{wrong} answers generated by GPT-4V.}
	\label{Figure12}
\end{figure*}

\begin{figure*}[t]
	\centering
	\includegraphics[width=0.8 \linewidth]{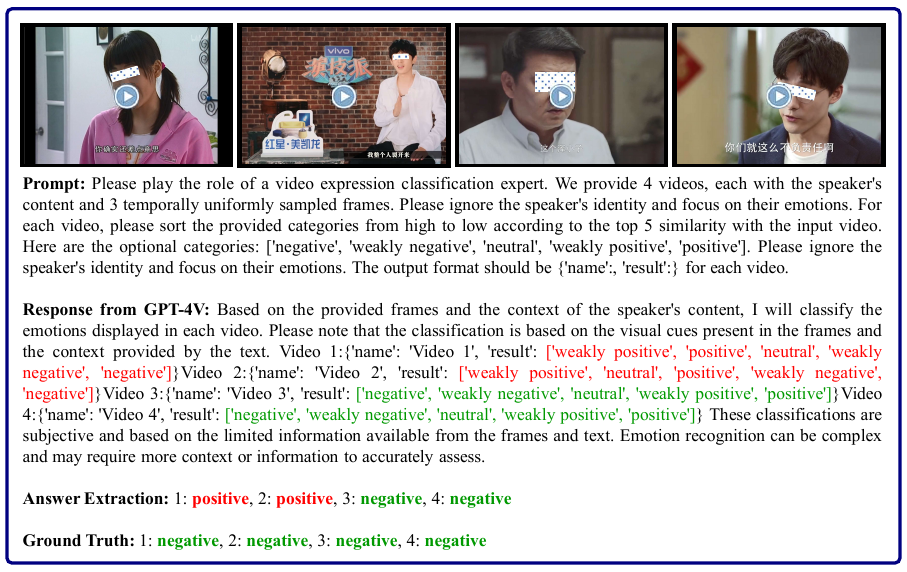}
	\caption{Case study for multimodal emotion recognition. We provide examples with \textcolor[rgb]{0,0.6,0}{correct} and \textcolor[rgb]{1,0,0}{wrong} answers generated by GPT-4V.}
	\label{Figure13}
\end{figure*}




% \end{thebibliography}

\end{document}